%% file: acl_latex.tex
\documentclass[11pt]{article}

\usepackage[preprint]{acl}
\usepackage{etoolbox}

\newtoggle{anonymise}
\togglefalse{anonymise}

\makeatletter
\@ifpackagewith{acl}{review}{
  \toggletrue{anonymise}
}{
  \togglefalse{anonymise}
}
\makeatother

\input{def}

\usepackage{times}
\usepackage{latexsym}

\usepackage[T1]{fontenc}

\usepackage[utf8]{inputenc}

\usepackage{microtype}

\usepackage{inconsolata}

\usepackage{graphicx}

\title{Who Wrote the Book? Detecting and Attributing LLM Ghostwriters}

\author{
  Anudeex Shetty$^{1}$,  Qiongkai Xu$^{1,2}$, Olga Ohrimenko$^{1}$, and Jey Han Lau$^{1}$ \\
  $^1${School of Computing and Information Systems, The University of Melbourne, Australia} \\
  $^2${School of Computing, FSE, Macquarie University, Australia} \\
  {\tt \{anudeex, oohrimenko, laujh\}@unimelb.edu.au,} \\
  {\tt qiongkai.xu@mq.edu.au }
}

\begin{document}
\maketitle
\begin{abstract}

\add{
Out-of-distribution (OOD) generalisation is essential for authorship attribution (AA), especially when involving new authors, yet this aspect remains insufficiently supported by existing AA datasets.
In this paper, we introduce \ourdataset, a dataset for LLM AA task, designed to test generalisation across multiple OOD dimensions, including unseen domains and authors. \ourdataset comprises long-form texts (avg. 49K words per book) generated by ten frontier LLMs. We find that existing methods
degrade substantially in unseen author scenarios --- dropping by as much as 70\% even under high-resource settings.
We propose \ourmethod, a novel lightweight and interpretable fingerprinting method that captures token-level transition patterns (\ie word rank and entropy) using a lightweight language model. Experiments show robustness of \ourmethod in OOD settings and  establish it as a strong baseline on  \ourdataset.\footnote{The dataset and code can be found at 
\iftoggle{anonymise}{
    \url{https://anonymous.4open.science/r/GhostWriteBench-TRACE}.
}{
    \url{https://github.com/anudeex/TRACE}.
}}
}

\end{abstract}

\section{Introduction}

Recently, a large number of AI detection datasets have been proposed  (see \reftab{tab:existing-datasets-overview}). However, several limitations remain. First, most datasets frame detection as binary human \versus AI classification \citep{he2024mgtbench,dugan-etal-2024-raid}, with only a few \citep{la2025openturingbench,uchendu2021turingbench} supporting the more challenging authorship attribution (AA) task (\ie multi-class model classification). Second, they rarely evaluate generalisation (\eg by including text generated by unseen generators or text from other domains).\footnote{\add{In this paper, we target specifically non-adversarial OOD generalisation (\eg unseen authors and domains) rather than adversarial robustness (\eg against obfuscation attacks \citep{dugan-etal-2024-raid}).}}  Finally, existing datasets focus on short texts, even though we are beginning to see the emergence of LLM-ghostwritten\footnote{Ghostwriting occurs when a writer (\aka \textit{ghostwriter}) produces content on someone's behalf without authorship credit. LLM-ghostwritten books are typically low-quality sham books designed to deceive customers and take sales away from legitimate books.} books  \citep{authorsguildDrivingSurge,auauthorsguild}.
In addition to these dataset limitations, existing attribution methods \citep{la2025openturingbench,guo2024detective,tyo2023valla,fabien-etal-2020-bertaa} typically rely on supervised training with known authors and exhibit limited generalisation across domains and unseen models.

We approach these challenges by (i) constructing a new dataset, \ourdataset, for AA with multiple OOD dimensions and (ii) proposing a novel fingerprinting approach, \ourmethod, that is lightweight and has better generalisation capabilities. \ourdataset supports two OOD dimensions: \oodOne and \oodTwo, which cover genre shift and unseen LLM author evaluation, respectively.  \ourdataset primarily focuses on long-form texts (\ie books) generated by current frontier LLMs (on average over 49K words per document) to reflect an emerging form of AI content generation: AI ghostwriting \citep{authorsguildDrivingSurge,auauthorsguild}. 
\add{Our experiments show that existing attribution methods struggle under the \oodTwo setting, with strong methods suffering drops of up to 90\% in \lessprolific and 70\% in \moreprolific training scenarios, highlighting limited generalisation in prior work.}

\begin{table*}[htp]
    \centering
    \resizebox{0.99\linewidth}{!}{
    \begin{tabular}{l c c c c c c r r}
    \toprule[1.5pt]
    \multirow{2}{*}{\textbf{Dataset}} & {\textbf{Task}} & \multicolumn{2}{c}{\textbf{LLM Generators}} & \multicolumn{2}{c}{\textbf{OOD}} & \textbf{Long} & \textbf{Avg.} & \textbf{Num.} \\
        \cmidrule(lr){3-4} \cmidrule(lr){5-6}
        & {\textbf{Type}} & {Num.} & {Frontier\textsuperscript{†}} & {Domain} & {Author} & {\textbf{Form}} & {\textbf{Length}} & {\textbf{Docs}} \\
    \midrule
        \textsc{TuringBench}$_{(\citeyear{uchendu2021turingbench})}$     & {Multi} & 20 & \xmark & \xmark & \xmark & \xmark & {<200} & {160K} \\
        \textsc{MGTBench}$_{(\citeyear{he2024mgtbench})}$        & {Binary} & 6 & \xmark  & \xmark & \xmark & \xmark & {<500} & {21K} \\
        \textsc{RAID}$_{(\citeyear{dugan-etal-2024-raid})}$            & {Binary} & 11 & \xmark & \cmark & \xmark & \xmark & {<250} & {500K} \\
        \textsc{M4GT-Bench}$_{(\citeyear{wang2024m4gt})}$      & {Multi} & 6 & \xmark  & \cmark & \xmark & \xmark & {<500} & {87K} \\
        \textsc{OpenTuring}$_{(\citeyear{la2025openturingbench})}$ & {Multi} & 7 & \xmark  & \cmark & \xmark & \xmark & {<500} & {497K} \\
        \midrule
        \textbf{\ourdataset}$_{\text{(Ours)}}$            & {Multi} & 10 & {\cmark} & {\cmark} & {\cmark} & {\cmark} & {49K} & {325} \\
    \bottomrule[1.5pt]
    \multicolumn{9}{@{}l}{\textsuperscript{†}{\footnotesize Frontier LLMs refer to recent state-of-the-art models (\eg \gptFive, \gptFour, \etc)}.} \\
    \vspace{-0.6em}
    \end{tabular}
    }
    \vspace{-1.25em}
    \caption{Overview of existing datasets for LLM-generated text attribution. 
    Most prior work focuses on binary classification rather than multi-class authorship attribution (AA) tasks.
    LLMs included in these benchmarks are largely outdated \add{(\ie they are small-scale, $7$B–$10$B parameters, models and/or not the latest frontier models)}, as indicated by the `Frontier' column. 
    While some studies include out-of-distribution (OOD) evaluation, this is rarely considered for AA tasks. `Author' OOD denotes open-set evaluation where authors unseen during training are evaluated. 
    `Long Form' refers to documents longer than $10$K words. 
    We report only AI-generated texts for `Avg. Length' (in words) and `Num. Docs'.
    }
    \label{tab:existing-datasets-overview}
\end{table*}

Next, we introduce a novel detection technique, \textbf{T}ransition-based \textbf{R}epresentation for \textbf{A}uthorship \textbf{C}lassification and \textbf{E}valuation (\ourmethod), for LLM attribution. \ourmethod first uses a (small) set of training texts generated by an LLM to create its fingerprint. At test time, to determine whether the test instance is generated by the LLM, we first create the test fingerprint and assess its similarity to the LLM's fingerprint. To produce the fingerprint for a text (during training or inference), we examine the transitions of token-level patterns (\eg rank of a word) estimated by another lightweight language model in the word sequence to create a compressed two-dimensional signature (see \reffig{fig:fingerprint-overview}). 
Our technique does not require white-box access and operates post hoc (and so works for both open- and closed-source LLMs). It requires minimal training data (\ie only a small set of LLM-generated texts), and is effective for detecting text from new domains or generated by unseen authors (LLM or human).

To summarise, our contributions are:
\begin{itemize}[itemsep=-1pt, topsep=3pt, leftmargin=10pt]
    \item We introduce \ourdataset, a new dataset for LLM authorship attribution that supports long-form book detection and explicitly tests generalisation across two OOD dimensions.
    \item We propose \ourmethod, a lightweight and interpretable fingerprinting technique for LLM authorship attribution. It achieves strong performance in-domain, OOD, and in \lessprolific training scenarios, establishing a strong baseline on \ourdataset.
\end{itemize}

\section{Related Work}

\begin{table*}[!htp]
    \centering
    \resizebox{0.9\linewidth}{!}{%
    \begin{tabular}{l S[table-format=2.0]
    S[table-format=2.1]  SSSSS}
        \toprule[1.5pt]
        \multirow{2}{*}{\textbf{Model}} 
            & \multicolumn{1}{c}{\textbf{Total}} 
            & {\textbf{Avg. \#}} 
            & \multicolumn{5}{c}{\textbf{Text Quality and Diversity}} \\
        \cmidrule(lr){4-8}
            & \multicolumn{1}{c}{\textbf{Books}} &
            \textbf{Words} & \multicolumn{1}{c}{PPL ($\downarrow$)} & \multicolumn{1}{c}{S-B ($\downarrow$)} & \multicolumn{1}{c}{S-R ($\downarrow$)} & \multicolumn{1}{c}{NGD ($\uparrow$)} & \multicolumn{1}{c}{CR ($\downarrow$)} \\
        \midrule
        \deepseek       & 42 & 33.4K & 21.62 & 0.001 & 8.22  & 2.29 & 2.96 \\
        \glm            & 32 & 28.4K & 33.06 & 0.002 & 7.69  & 2.46 & 2.90 \\
        \kimi           & 37 & 37.1K & 27.87 & 0.002 & 8.76  & 2.24 & 2.81 \\
        \qwen           & 38 & 67.1K & 21.30 & 0.004 & 10.04 & 2.18 & 2.72 \\
        \qwenMax        & 34 & 41.2K & 24.96 & 0.003 & 9.01  & 2.29 & 2.73 \\
        \claude         & 24 & 47.1K & 18.89 & 0.005 & 8.38  & 2.37 & 2.88 \\
        \geminiFlash    & 30 & 20.5K & 24.18 & 0.003 & 7.68  & 2.37 & 2.88 \\
        \geminiPro      & 27 & 88.1K & 22.10 & 0.004 & 9.89  & 2.06 & 2.71 \\
        \gptFour        & 35 & 66.1K & 40.59 & 0.002 & 8.53  & 2.44 & 2.58 \\
        \gptFive        & 26 & 69.3K & 31.67 & 0.005 & 9.04  & 2.33 & 2.64 \\
        \midrule
        Human & \multicolumn{1}{c}{100} & 88.0K & 31.36 & 0.001 & 10.25 & 2.14 & 2.65 \\
        \bottomrule[1.5pt]
    \end{tabular}
    }
    \caption{\ourdataset statistics: total books (genre splits in \refapptab{tab:detailed-dataset-statistics}), average number of words in \doc, and text quality and diversity metrics for each LLM (and 100 random human books). Metrics include perplexity (PPL), self-BLEU (S-B), repetition (S-R), $n$-gram diversity (NGD), and compression ratio (CR) (more details in \refapp{sec:quantiative-analysis-dataset}). Arrows ($\uparrow$; $\downarrow$) indicate whether higher or lower values are better.
    }
    \label{tab:stats-quantative-eval-dataset}
\end{table*}

\subsection{Authorship Attribution}
Authorship attribution (AA) is a long-standing field \citep{mendenhall1887characteristic,mosteller1963inference} that has grown with the rise of digital texts and advances in machine learning \citep{tyo2023valla,stamatatos2009survey,neal2017surveying}. It is usually formulated as a multi-class classification problem, with each class representing an author. Current SOTA methods rely on neural networks, often using Transformer-based backbones fine-tuned on domain-specific corpora \citep{tyo2023valla,fabien-etal-2020-bertaa}. Despite their effectiveness, these approaches are resource-intensive, require careful dataset curation, generalise poorly to new domains, and struggle with long documents due to standard Transformer \citep{vaswani2017attention} context limits (\eg 512 tokens).

\subsection{LLM-Generated Text}
As LLMs are increasingly used to generate misinformation, provenance and veracity have become pressing \citep{kumarage2024survey,huang2025authorship}. Detecting LLM-generated text is non-trivial: humans often perform only slightly above chance \citep{dou-etal-2022-gpt,uchendu2021turingbench}. Most prior work \citep{mitchell2023detectgpt,yang2024dnagpt,wu2025survey} focuses on binary (human \versus machine) classification, even though attribution to a specific model is important for accountability. Existing SOTA methods \citep{la2025openturingbench,guo2024detective} largely mirror traditional AA approaches, inheriting limitations such as poor generalisation and lack of interpretability. Some works have explored cross-domain evaluation \citep{dugan-etal-2024-raid,he2024mgtbench,wang2024m4gt}. However, they consider only a limited set of authors and exclude long-form LLM-generated text. Most assume a closed set of authors, whereas practical attribution often occurs in open-set scenarios \citep{stolerman2014breaking,huang2025authorship}, \add{which is the setting that \ourdataset assumes.}

LLMs have also fostered ghostwriting, producing content on someone's behalf without authorship credit, contributing to low-quality or sham books impacting authors \citep{authorsguildDrivingSurge,auauthorsguild}. Motivated by these developments and methodological limitations, our work focuses on improving the generalisation of attribution techniques for LLM-generated long-form documents such as books.\footnote{Although ghostwriting can occur in short texts, it mainly affects long-form documents, making books a practical focus.}

\section{\ourdataset Dataset}

In this paper, we focus on the task of LLM AA, \ie determining which LLM generated a given document. 
As shown in \reftab{tab:existing-datasets-overview}, most existing datasets of LLM-generated text focus on constrained settings, \ie (i) framing the task as binary detection, (ii) comprising outdated models, (iii) without evaluating OOD generalisation, and (iv) focusing on short texts.  Motivated by these limitations, we construct a new dataset, \ourdataset, covering ten recent frontier LLMs (model details in \refapptab{tab:model-card}) and comprising books (both fiction and non-fiction) as long-form documents\footnote{Our definition of long-form is document with more than $10$K words.} supporting multiple OOD dimensions.

\subsection{\Doc Generation Pipeline}
\label{sec:novel-gen}

Humans typically write a \doc through a hierarchical process involving planning, drafting, and revision \citep{sun2022summarize,lee2025navigating,hayes2016identifying}. We mimic this process by designing a \doc generation pipeline with incremental steps: outlining, summarisation, and long-text expansion, following prior works \citep{yao2019plan,wang2025towards,alhussain2021automatic}. 

For each book, we first generate a detailed outline conditioned on some constraints (\eg era and genre); this outline produces the narrative structure, plot, and characters. 
And then iteratively over multiple turns, we instruct the LLM to produce the next text segment, conditioned on the outline, the previous segment, and the running summary (prompts and more details in \refapp{sec:novel-gen-prompts}). We provide real, human-authored texts (sourced from Project Gutenberg\footnote{\url{https://www.gutenberg.org/}.}) as few-shot reference examples to guide the LLM generation to maximise generation quality (constraints such as era and genre are therefore drawn from these Gutenberg books). This pipeline ensures that narrative coherence and discourse structure are preserved, enabling the generation of full-length books.\footnote{\Doc generation statistics reported in \refapptab{tab:generation-stat-dataset}.}
\add{Importantly, all books were generated using the same generation pipeline across all LLMs, ensuring no pipeline-specific bias.  While some model-specific generation biases are inevitable, we try to reduce the likelihood of dataset artefacts through careful data cleaning (\refsec{sec:data-cleaning}).}
\reftab{tab:stats-quantative-eval-dataset} provides various statistics for these LLM-generated books; we will return to discuss more on how we evaluate quality in \refsec{sec:dataset-analysis}.

\subsection{OOD Dimensions}
\label{sec:ood-dims}

In addition to standard in-distribution (ID) evaluation, 
 the \ourdataset dataset examines two types of OOD generalisation: 
 (i) \textbf{\oodOne}, to test generalisation to texts written in unseen genres; and (ii) \textbf{\oodTwo}, to test generalisation to texts generated by unseen LLMs.
We construct per-author domain splits, where book genres are partitioned into ID and OOD sets for each author.\footnote{We ensure there is no genre overlap between the ID and \oodOne partition per author; see Appendix \reftab{tab:detailed-dataset-statistics}.} This creates a more challenging setting than using a single split shared across all authors, as commonly used in existing datasets (see \refapptab{tab:detailed-dataset-statistics} for genre splits). We use the genres defined by \citet{momen-etal-2025-filling} for Gutenberg books (see \reftab{tab:genres-info} for the list of genres\footnote{Recall that we use Gutenberg books as few-shot reference examples for the LLM. As such, the genres for the LLM-generated books are also grounded in Gutenberg genres.}).
In practice, a Gutenberg book often consists of multiple genres (\eg \textsf{LF} (Literature \& Fiction) $+$ \textsf{ST} (Science \& Technology), and many of the LLM-generated books are also multi-genre.

For \oodTwo, we create ten splits by holding out one LLM at a time (we have ten LLMs) \addThree{and report average performance across these splits}. That is, for a given split, the training partition contains books from nine LLMs and the test partition contains books from the held-out LLM \addThree{to ensure there is no development-set leakage during thresholding}.

To understand the impact of training data, within a particular evaluation setting (ID, \oodOne, \oodTwo), \ourdataset also provides two training scenarios: \textbf{\moreprolific} and \textbf{\lessprolific}, where the former has a large number (10--35) and the latter a small number of training books (3--5) text--label pairs per LLM.\footnote{To ensure robustness of results, we create three random splits for each subset; see \refapptab{tab:prolificity-splits}.}

\subsection{Dataset Analysis}
\label{sec:dataset-analysis}

\begin{figure*}[!htp]
    \centering
    \includegraphics[width=0.9\linewidth]{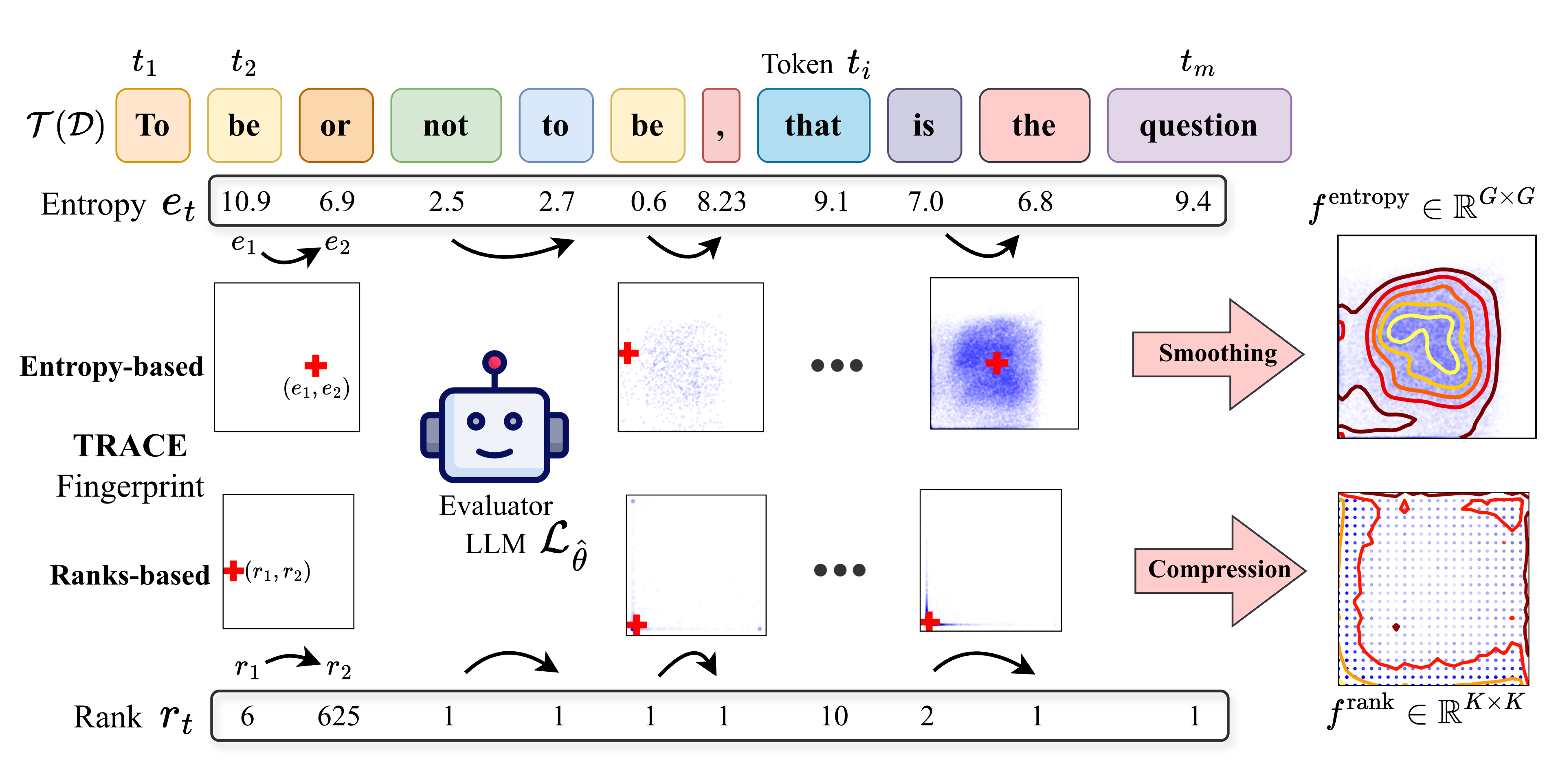}
    \vspace{-0.75em}
    \caption{
    Overview of \ourmethod fingerprint construction. We model transitions of consecutive token scores (using another lightweight evaluator language model; \gpt used in illustration). The process is shown for \addThree{both Rank-based (\S\ref{sec:rank-based-fingerprint}) and} Entropy-based (\S\ref{sec:entropy-based-fingerprint}) fingerprint variant. We construct a pool of reference fingerprints from an LLM's training texts. At inference time, attribution is a straightforward comparison between the test fingerprint and the LLM reference fingerprint. 
    }
    \label{fig:fingerprint-overview}
\end{figure*}

\paragraph{Data Cleaning.} 
\label{sec:data-cleaning}
We sanitise the LLM-generated books by removing non-ASCII and smart characters, along with common LLM markers (\eg END, START OF SEGMENT). As observed by \citet{momen-etal-2025-filling}, the start and end pages of the books often contain identifier markers (such as author, title, year, and others). Therefore, we remove $1.5$K tokens from the beginning and end of the text to prevent potential shortcuts by attribution methods. We also conduct manual inspection on a subset and perform $n$-gram analysis across the corpus to check for frequent boilerplate text (titles, generation markers, \etc) to filter any additional shortcuts.

\paragraph {Quality Analysis.} We evaluate text quality using several metrics in \reftab{tab:stats-quantative-eval-dataset}. We measure perplexity (PPL; \citet{jm3}) of generated texts for different LLMs, a standard measure of language understanding. 
We then measure several lexical diversity metrics, comprising both inter- and intra-document measures. We compute self-\textsc{BLEU} (S-B; \citet{zhu2018texygen}), a homogenisation score over pairs of books generated by the LLM (since we lack references), which is commonly used in writing tasks to ensure diversity \citep{padmakumar2024does}. It is common for LLMs to repeat text in long-form generation \citep{bai2025longwriter}; hence, we evaluate self-repetition (S-R; \citet{salkar-etal-2022-self}), which captures this repetition tendency. Similarly, we measure the $n$-gram diversity score (NGD; \citep{meister2023locally}), which measures the number of unique $n$-grams in the texts generated by LLMs. Finally, a non-$n$-gram-based metric, compression ratio (CR; \citep{shaib-etal-2025-standardizing}) evaluates compressibility (higher values suggest redundancy). Across all these diversity metrics, we observe that all LLMs in \ourdataset have values similar to those of human books, quantitatively indicating comparable text quality.

In addition to quantitative analysis, we conduct a small-scale human evaluation on a subset of books (more details in \refapp{sec:qualitative-analysis-dataset}) with ten annotators. We assess quality across multiple dimensions, following prior work \citep{xia2025storywriter,chhun2022human}. 
\refapptab{tab:human_eval_llms} reports the average annotator scores across different quality dimensions (see \refapp{sec:qualitative-analysis-dataset} for definitions), both at the passage level and overall for each \doc. Overall, we observe strong scores ($>4$; out of 5) for relevance, coherence, fluency, and diversity, complementing the above text quality and diversity analysis. Scores for other subjective dimensions (engagement, empathy, surprise, and complexity) are also generally positive ($>3.3$; out of 5). These results indicate that the text in \ourdataset exhibits strong relevance and is of overall {good quality}.

\section{\ourmethod Fingerprint}
\label{sec:fingerprint}

We next describe \textbf{T}ransition-based \textbf{R}epresentation for \textbf{A}uthorship \textbf{C}lassification and \textbf{E}valuation (\ourmethod), a novel fingerprinting method for LLM attribution. At training time, we first construct reference fingerprints for each LLM (if an LLM has multiple books, we create one fingerprint for each book). Given a test sample, we then construct its fingerprint and determine whether it matches one of the reference fingerprints.\footnote{Rather than performing majority voting over reference fingerprints, we retrieve the most similar reference fingerprint.}

As different LLMs may have different preferences when it comes to word selection during generation, our idea is to examine token-level transitions \addTwo{(as shown in \reffig{fig:fingerprint-overview})}, such as word rank (\refsec{sec:rank-based-fingerprint}) and entropy (\refsec{sec:entropy-based-fingerprint}), as computed by a separate, lightweight \textit{evaluator} language model. \add{
Although capturing word preferences via a separate evaluator model is related to existing distributional heuristics (\eg perplexity; \citet{wu2025survey,huang2025authorship}), \ourmethod is fundamentally different in that it models \textit{transitions} between consecutive tokens rather than static distributions, producing more discriminative signatures (as seen in Appendix~Figures~\ref{fig:entropy-ref-fingerprints} and~\ref{fig:ranks-ref-fingerprints}). To the best of our knowledge, modelling token preference transitions as fingerprints for multi-class LLM attribution has not been explored in prior work.}

Our fingerprint \ourmethod can be applied to both open- and closed-source LLMs, since it does not require white-box access. It is lightweight and adaptable, and produces an interpretable 2D signature (\addTwo{more in \refsec{sec:qualitative-trace}}).
We next discuss these two variants of \ourmethod: Rank-based and Entropy-based.

\subsection{Rank-Based}
\label{sec:rank-based-fingerprint}

Formally, given a document (or book in our case) $\mathcal{D}$ represented as a token sequence $\langle t_{1}, t_{2}, \dots, t_{m}\rangle$ tokenised by a tokenizer $\mathcal{T}$ with vocabulary $\mathcal{V}$.
For each token $t_i \in \langle t_1, t_2, \dots, t_m \rangle$, we compute its rank under an evaluator language model $\mathcal{L}_{\hat{\theta}}$ with context $t_{<i}$ as follows:
\add{\small
\begin{equation} \notag
r_i
=
\left|
\left\{
v \in \mathcal{V} :  \mathbb{P}(t=v \mid t_{<i}; \hat{\theta}) \ge  \mathbb{P}(t=t_i \mid t_{<i}; \hat{\theta})
\right\}
\right|.
\end{equation}}

\begin{figure*}[!htp]
    \centering
    \includegraphics[width=0.8\linewidth]{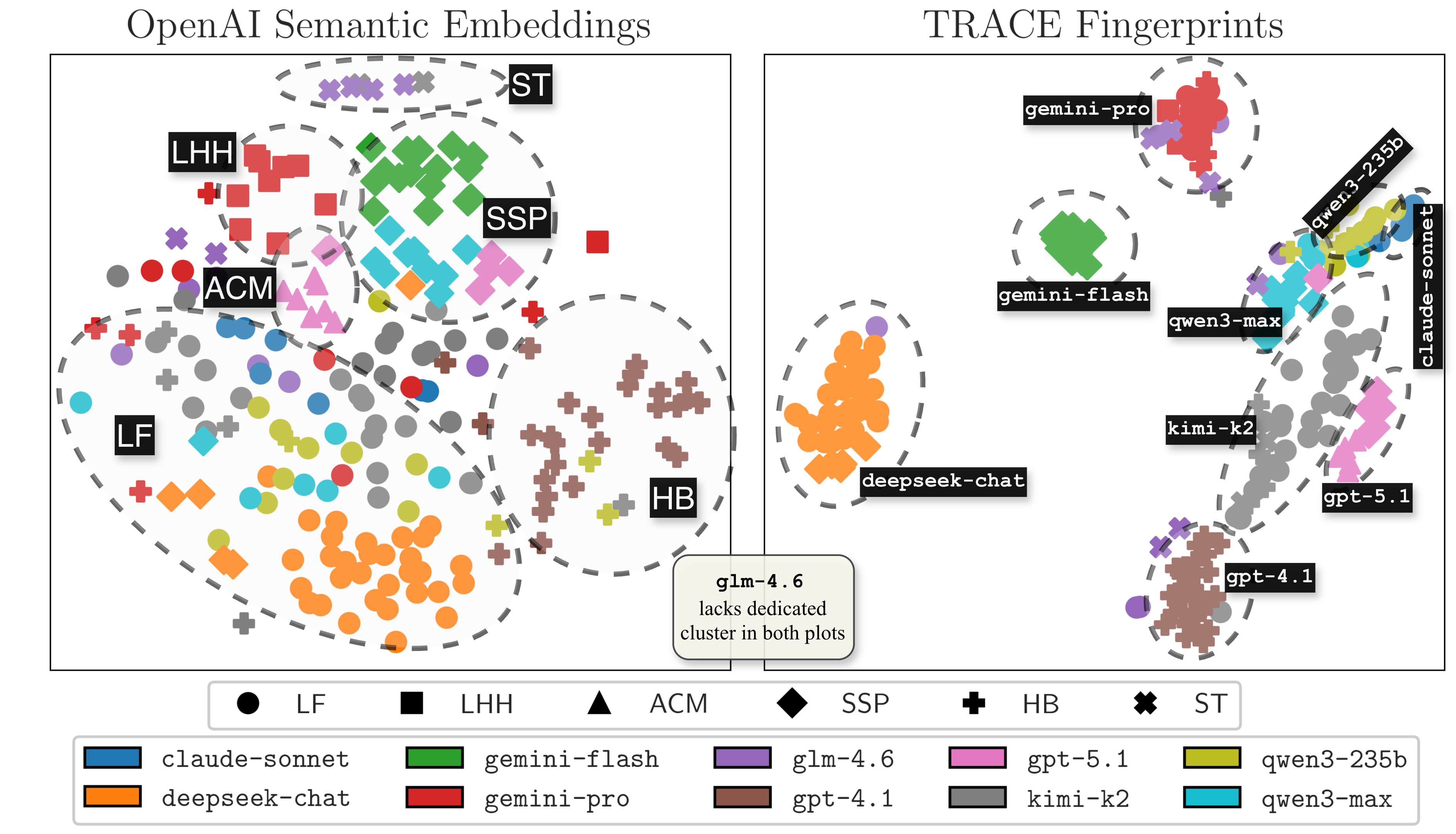}
    \caption{\add{t-SNE visualisation comparing semantic embeddings (\texttt{openai/text-embedding-3-large}; left) and our Entropy-based \ourmethod fingerprints (right) on single-genre books in \ourdataset. Marker shape denotes genre and colour denotes the generating LLM. The left plot shows content-based clustering by genre, while the right plot presents clusters by LLM across genres, indicating that \ourmethod captures model-specific stylistic patterns rather than genre-specific content features.
    }}
    \label{fig:umap-semantic-vs-trace}
\end{figure*}

The raw fingerprint  $\hat{f}^{\operatorname{rank}} \in \mathbb{R}^{|\mathcal{V}|\times|\mathcal{V}|}$ captures the counts of $(r_{i-1}, r_i)$ (left side in \reffig{fig:fingerprint-overview}).
This fingerprint matrix is very large and sparse, and so we compress it by binning ranks into equal probability clusters (as detailed in \refappalg{algo:rank-transition-fingerprint}), giving the final fingerprint $f^{\operatorname{rank}} \in \mathbb{R}^{K\times K}$ where $K \ll |\mathcal{V}|$.

We apply this fingerprint construction process for each LLM using their training books (if an LLM has multiple books, one fingerprint is created for each book) to create reference fingerprints for each LLM. Then, for a test text, we construct a test fingerprint following the same process above. To determine which LLM generated the test text, we identify the reference fingerprint that is most similar to the test fingerprint. 

To measure similarity between two fingerprints, which can be interpreted as discrete probability distributions, we use the Jensen-Shannon (JS) distance (equation given in \refapp{sec:sim-metrics}).

\subsection{Entropy-Based}
\label{sec:entropy-based-fingerprint}
Entropy
has been used in NLP to capture various properties of natural language \citep{genzel2002entropy,jaeger2010redundancy}.
Like word ranks, LLMs may exhibit unique entropy transition patterns and so constitute a potential signal for creating the fingerprint. We provide a qualitative example in \refappfig{fig:example-fingperprints}, illustrating the generalisation of Entropy-based fingerprints.

Formally, the entropy of a token $t_i$ similar to setup in Rank-based is given by:
\begin{equation} \notag
    e_i = -\sum_{v\in\mathcal{V}} \mathbb{P}(t=v \mid t_{<i}; \hat{\theta}) \cdot \log \mathbb{P}(t=v \mid t_{<i}; \hat{\theta}).
\end{equation}
Unlike word ranks, the resulting fingerprint is continuous rather than discrete (intuitively the fingerprint is like a scatter plot). To smooth it and produce a density map (of dimension $G$), 
 we use kernel density estimation  (detailed in \refappalg{algo:entropy-estimation}) to obtain $f^{\operatorname{entropy}} \in \mathbb{R}^{G\times G}$ where $G$ is the grid size of the density map.
For fingerprint matching, in addition to JS distance (as used for ranks), we experiment with another similarity metric, norm-mean, which is computed by comparing the angle between surface normals (equation given  in \refapp{sec:sim-metrics}).

\addTwo{
\subsection{Qualitative nature of \ourmethod fingerprints}
\label{sec:qualitative-trace}
}

\addTwo{As illustrated in \reffig{fig:fingerprint-overview}, we construct the fingerprint by modelling transitions between consecutive tokens. Each axis of the resulting 2D matrix represents the rank (or entropy) of consecutive token transitions as estimated by an evaluator language model, and the density in a given region reflects how frequently an LLM produces that transition type. For instance, high density in the bottom-left indicates a preference for high-probability tokens following other high-probability tokens (a more predictable generation style), while density in the top-right reflects a tendency toward more surprising token choices. Different LLMs produce visually distinct fingerprints; for example, GPT models exhibit a more circular structure (for Entropy-based; ~\refappfig{fig:entropy-ref-fingerprints}) than Gemini models, which tend toward a squarer pattern.} %

\addTwo{We hypothesise that \ourmethod generalises because it relies on token-level transition patterns computed by an independent evaluator language model, rather than direct supervision on attribution labels (as done in prior methods). This is supported by the preliminary analysis in \reffig{fig:umap-semantic-vs-trace}, where the \ourmethod's signature for both ID and OOD books from the same LLM cluster together, suggesting that \ourmethod captures LLM-specific stylistic characteristics rather than content; in contrast, semantic embeddings are genre-clustered.} \addTwo{We next study this empirically with experiments on \ourdataset.}

\vspace{0.75em}
\section{Experimental Setup}

\begin{table*}[!htp]
    \centering
    \resizebox{0.95\textwidth}{!}{
    \begin{tabular}{l c c c c c c}
        \toprule[1.5pt]
        \multirow{2}{*}{\textbf{Method}} & \multicolumn{3}{c}{\textbf{Low Resource}} & \multicolumn{3}{c}{\textbf{High Resource}} \\
        \cmidrule(lr){2-4} \cmidrule(lr){5-7}
         & ID & \oodOne & \oodTwo & ID & \oodOne & \oodTwo \\
        \midrule
        \textbf{\rank} & $0.00 \pm 0.00$ & $0.00 \pm 0.00$ & $0.40 \pm 0.19$ & $0.27 \pm 0.08$ & $0.25 \pm 0.06$ & $0.38 \pm 0.21$ \\
        \textbf{\entropy} &  $0.00 \pm 0.00$ & $0.00 \pm 0.00$ & $0.20 \pm 0.20$ & $0.33 \pm 0.06$ & $0.35 \pm 0.04$ & $0.10 \pm 0.08$ \\
        \textbf{\gltr} &  $0.00 \pm 0.00$ & $0.00 \pm 0.00$ & $0.44 \pm 0.17$ & $0.30 \pm 0.02$ & $0.31 \pm 0.02$ & $0.28 \pm 0.03$ \\
        \midrule
        \textbf{\ngram} & $0.40 \pm 0.16$ & $0.20 \pm 0.16$ & $0.36 \pm 0.31$ & \underline{$0.96 \pm 0.01$} & $\bm{0.92 \pm 0.04}$ & $0.28 \pm 0.35$ \\
        \textbf{\bert} & $0.44 \pm 0.17$ & $0.27 \pm 0.19$ & $0.04 \pm 0.06$ & $0.96 \pm 0.04$ & \underline{$0.90 \pm 0.05$} & $0.29 \pm 0.05$ \\
        \textbf{\topformer} & $0.51 \pm 0.28$ & $0.24 \pm 0.17$ & $0.29 \pm 0.17$ & $0.94 \pm 0.04$ & $0.88 \pm 0.04$ & $0.46 \pm 0.27$ \\
        \textbf{\detective} & $0.53 \pm 0.09$ & \underline{$0.27 \pm 0.19$} & $0.40 \pm 0.20$ & $0.95 \pm 0.03$ & $0.85 \pm 0.11$ & \underline{$0.49 \pm 0.22$} \\
        \midrule
        \textbf{\ourmethod (Ours)} && \\
        \ \ {Ranks-JS} & $0.40 \pm 0.24$ & $0.22 \pm 0.19$ & $0.18 \pm 0.03$ & $0.78 \pm 0.06$ & $0.66 \pm 0.12$ & $0.20 \pm 0.08$ \\
        \ \  {Entropy-JS} & \underline{$0.60 \pm 0.16$} & $0.23 \pm 0.12$ & \underline{$0.49 \pm 0.14$} & $0.91 \pm 0.06$ & $0.82 \pm 0.08$ & $\bm{0.58 \pm 0.20}$ \\
        \ \ {Entropy-Norm} & $\bm{0.67 \pm 0.19}$ & $\bm{0.49 \pm 0.11}$ & $\bm{0.51 \pm 0.17}$ & $\bm{0.96 \pm 0.02}$ & $0.82 \pm 0.10$ & $0.37 \pm 0.03$ \\
        \bottomrule[1.5pt]
    \end{tabular}}
    \caption{Results for attribution techniques on \ourdataset under \lessprolific (left) and \moreprolific (right) training settings for different OOD dimensions (\oodOne and \oodTwo). Reported scores are \fscore (mean $\pm$ std for three random splits; see \refapptab{tab:prolificity-splits}); \textbf{bold} denotes the best result and \underline{underline} the second best. \add{Overall, \ourmethod has strong performance with several existing methods showing limited generalisation.}
    }
    \label{tab:main-results}
\end{table*}

\paragraph{Baselines.}

We consider techniques from different areas and categories as baselines. From metric-based methods, we evaluate \rank, \entropy, and \gltr \citep{gehrmann2019gltr}. Although originally designed for binary detection,  we follow \citet{la2025openturingbench} to adapt them for multi-class classification (by learning a threshold for each LLM). 

\addTwo{
For model-based (supervised) methods, we consider \ngram \citep{koppel2004authorship}, \bert \citep{fabien-etal-2020-bertaa}, \topformer \citep{uchendu2024topformer}, and \detective \citep{guo2024detective}, covering both non-neural and neural (including different backbones, representations, and learning paradigms) approaches. \citet{tyo2023valla} found very strong performance for \ngram and \bert for traditional human AA tasks. \topformer is a \textsc{RoBERTa}-based model fine-tuned with a topological data analysis (TDA) layer specifically designed for the AA task, capturing distinctive writing styles. While \detective is a SOTA multi-level contrastive learning-based detector for AI-generated text with implicit author representation learning. Finally, while there are recent LLM-based detection methods \citep{binoculars,yu-etal-2024-text,verma-etal-2024-ghostbuster}, most are formulated for binary human-vs-machine detection and are therefore not directly applicable to the LLM attribution setting considered in this work. More details regarding baselines are provided in \refapp{sec:baselines}.
}

\paragraph{Experimental Setting.}
We use \gpt (smallest version; $124$M parameters) \citep{radford2019language}
as the independent evaluator language model to estimate word ranks and entropy for creating the fingerprints.
For the Rank-based fingerprint, we use a cluster size ($K$)
of 50. For the Entropy-based fingerprint, we use a grid size ($G$) of 50. 
All hyperparameters are selected based on the development set \addThree{(which is additional two ID genre books per LLM)}. Additional experimental details in \refapp{sec:app-exp-details}.

\paragraph{Evaluation.} 
AA techniques are typically deployed in open-set settings, where the author of a test text may not appear among those seen during training (\ourdataset captures this with \oodTwo evaluation). In such cases, a reliable technique should be well-calibrated, assigning high confidence to known authors and low confidence to unseen ones (\ie rejection) \citep{scheirer2012toward,guo2017calibration}. To achieve this, we calibrate all methods (baselines and our methods) on the development set to determine a confidence threshold that best balances rejection and attribution performance. All thresholds for different configurations can be found in \refapptab{tab:thresholds-config}. During evaluation, we apply this threshold across all scenarios: ID, \oodOne, and \oodTwo for fair comparison (in a practical application, we would not know whether a test text is ID or OOD, and so this thresholding would have to be applied at all times). For texts from known authors, predictions below the threshold are treated as false negatives (missed attributions); for unseen authors, such predictions correspond to correct rejection. 
AA performance is evaluated using the \fscore metric, which assigns equal weight to all LLMs \citep{uchendu-etal-2020-authorship,tyo2023valla}.
\add{As discussed in \refsec{sec:fingerprint}, we use JS distance and norm-mean similarity metrics (more in \refapp{sec:sim-metrics}) for comparing \ourmethod fingerprints.}

\section{Experimental Results}

\subsection{Main Results}
\reftab{tab:main-results} presents the main results on \ourdataset, comparing the \ourmethod fingerprint against baselines. 
We report results separately for \lessprolific and \moreprolific authors (recall from \refsec{sec:ood-dims} that these settings control the amount of training books available per LLM); combined results can be found in \refapptab{tab:all_authors-main-results}.
Although many baselines perform well in the ID setting, they experience substantial performance degradation \add{(as high as 90\% in \lessprolific and 70\% in \moreprolific training settings)} in our OOD setting, showing that these methods would not generalise well in a real-world application.
In the more challenging scenarios: \lessprolific and the two \oodOne and \oodTwo settings, \ourmethod consistently outperforms both model- and metric-based baselines. A qualitative example is shown in \refappfig{fig:example-fingperprints}.
That said, for some settings with more training data (\moreprolific), a few baselines marginally outperform our method. \add{For completeness, we also report top-$k$ accuracy results in \refapptab{tab:all_authors-top-k-results} and without any thresholding (for ID and \oodOne) in \refapptab{table:w/o-thresholding-results}, and note that
the overall trend is the same as the main results (\reftab{tab:main-results}).}

\begin{table}[t]
    \centering
    \resizebox{0.95\linewidth}{!}{
    \begin{tabular}{l cc}
    \toprule[1.5pt]
        \multirow{2}{*}{\textbf{Method}} & \multicolumn{2}{c}{\textbf{\oodTwo}} \\
        \cmidrule(lr){2-3}
            & LLM & Human \\
        \midrule
        \textbf{\ngram }& $0.32 \pm 0.33$ & \underline{$0.48 \pm 0.29$} \\
        \textbf{\detective} & \underline{$0.45 \pm 0.15$} & $0.16 \pm 0.07$ \\
        \midrule 
        \textbf{\ourmethod  (Ours)} \\
        \ \ Ranks-JS & $0.19 \pm 0.03$ & $0.11 \pm 0.01$ \\
        \ \ Entropy-JS & $\bm{0.54 \pm 0.15}$ & $\bm{1.00 \pm 0.00}$ \\
        \ \ Entropy-Norm & {$0.44 \pm 0.09$} & $\bm{1.00 \pm 0.00}$ \\
        \bottomrule[1.5pt]
    \end{tabular}
    }
    \caption{Performance on the \oodTwo variant consisting of unseen human authors. \add{\ourmethod achieves near-perfect performance, demonstrating its ability to distinguish human-written \versus LLM-generated texts.}}
    \label{tab:unseen-human}
\end{table}

\paragraph{Unseen Human Texts.} 
\label{sec:unseen-human-results}
We also evaluate a variant of the \oodTwo setting, focusing on unseen \textit{human} authors. We randomly sample Gutenberg books by 10 human authors and measure how well our method can identify these unseen human texts, reflecting the common binary classification task of LLM-generated \versus human-written. As shown in \reftab{tab:unseen-human}, the Entropy-based variants of \ourmethod achieve perfect performance in distinguishing human authors from LLM authors. These results indicate that the fingerprints of human authors are \textit{very} different to those of LLM authors \add{(also supported by visualisations in \refappfig{fig:unseen-humans-entropy-ref-fingerprints})}, suggesting that \ourmethod can also be an effective detector in the more traditional binary classification setup.

\paragraph{Error Analysis.}
We examine the prediction confusion matrices (see \refappfig{fig:conf-matrix-OOD-results}) for the top-performing methods: \ngram, \detective, and the Rank- and Entropy-based \ourmethod variants. We observe that baselines tend to confuse similar LLMs (\ie those belonging to the same model family). This pattern is more pronounced in \refapptab{tab:family-OOD-results}, where we assess  \oodTwo in a  different complementary manner: if the predicted class for the unseen LLM belongs to the same family (\eg unseen \gptFour texts classified as \gptFive), we treat that as a correct prediction. Models such as \bert, \topformer, and \detective show a performance gain, indicating that they lack the specificity to identify a particular LLM and tend to confuse it with models from the same family. %

\subsection{Ablation Study}

\paragraph{Impact of Evaluator Language Model.}
We investigate the impact of using different evaluator language models ($\mathcal{L}_{\hat{\theta}}$), which form the main component of \ourmethod. These models are used to compute scores (rank and entropy) for each token, as defined in \refsec{sec:fingerprint}. We evaluate two other 1B models, \olmo and \gemma,\footnote{Exact model versions are \texttt{allenai/OLMo-2-0425-1B} and \texttt{google/gemma-3-1b-it}.} which differ in architecture, type, and size compared to \gpt used in  our experiments. We observe \addThree{similar performance across different evaluator language models in the majority of settings} (see \refapptab{tab:eval-llms-all}). This suggests that \ourmethod exhibits considerable robustness to the choice of evaluator language model.

\addTwo{\paragraph{Possible Data Contamination.} All classifiers, detectors, and evaluators used in our experiments---including \ourmethod and the baseline methods---are based on \gpt or \textsc{BERT}/\textsc{RoBERTa} architectures, which were not pre-trained on the Gutenberg corpus. For the frontier models serving as generators, there may be potential data contamination due to their pre-training data, since snippets from Gutenberg books are provided as few-shot references during \ourdataset \doc generation. Importantly, this affects only the generation process and not the evaluation stage. Consistent with this, we observe no significant performance differences in \ourmethod when using alternative evaluator language models (\eg \gemma, which likely includes Gutenberg data in their pre-training).}

Further ablation studies investigating different components of \ourmethod (size, order, and similarity metrics), numbers of tokens, and other hyperparameters can be found in \refapp{sec:app-ablation-studies}.

\section{Conclusion}
We propose \ourdataset, an authorship attribution dataset for long-form LLM texts. To study OOD generalisation, \ourdataset examines two OOD settings: \oodOne and \oodTwo. \add{Our experiments reveal that existing attribution methods show limited generalisability.}
We also introduce \ourmethod, a novel, interpretable fingerprint method for LLM attribution.  \ourmethod uses token-level transition patterns, such as word ranks and entropy, to create text fingerprints. Our experiments demonstrate the effectiveness of \ourmethod in ID, OOD, and limited training data scenarios in \ourdataset. \add{Our research provides a new benchmark that explicitly tests for generalisation and establishes a new strong baseline,} paving the way for future research directions in LLM text forensics. %

\section*{Limitations}
\ourdataset is limited to English texts only; we plan to extend this dataset with cross-lingual OOD dimensions in future work. It would be interesting to study the effectiveness of \ourmethod in an OOD-language setting.
Similarly, there are other OOD dimensions, such as time, human authors, and others, which are natural extensions of \ourdataset.
Although the books generated by our pipeline are of good quality, we acknowledge that \doc generation and related aspects, such as creativity, self-reflection, rewriting, \etc, constitute research problems in their own right and fall outside the scope of this work.
\add{Adversarial robustness (\eg paraphrasing or style-transfer attacks) is an orthogonal generalisation axis not studied in this work. That said, since \ourmethod is not a supervised method, we expect it to be less prone to such obfuscation attacks; a thorough evaluation remains an important direction for future work.}
Adapting \ourmethod for fewer (<10K) tokens remains an open problem.
\add{For \bert, \topformer, and \detective, we use chunking to handle long texts, following prior work. Alternatives such as filtering chunks by confidence or using overlapping windows are interesting avenues for future work.}
We also focus only on a single-authorship setup. A natural next step would be to study multi-authorship (\ie a piece of text written by multiple authors), which is more common in long-form documents that typically involve collaborative writing.

\section*{Ethics Statement}
\ourdataset contains LLM-generated books across various genres. The content of these books may include writing or plot elements that reflect undesirable values. That said, all LLMs used incorporate built-in guardrails, so the likelihood of such issues should be minimal. This is further supported by our small-scale quality annotation, where no annotators identified any issues.

\iftoggle{anonymise}{}{
\section*{Acknowledgements}

\addThree{We appreciate the valuable feedback from all anonymous reviewers.
We thank annotators Liuliu Chen, Ming-Bin Chen, Edoardo De Duro, Matteo Greco, Matteo Guida, Jemima Kang, Demian Inostroza, and Rui Xing for their time and support in data annotation.
}

\addThree{This research was supported by The University of Melbourne's Research Computing Services and the Petascale Campus Initiative.
A.\ Shetty was supported by the Commonwealth through an Australian Government Research Training Program Scholarship (DOI: \url{https://doi.org/10.82133/C42F-K220}), Computing and Information Systems PhD Scholarship, and the Avashya Foundation. A.\ Shetty's conference travel was partially supported by the Engineering and IT Conference Travel Grant. J.H.\ Lau was supported by
the Australian Research Council under Grant LP210200917 and DP240101006.
}
}

\bibliography{custom}

\clearpage
\appendix
\section*{Appendix}

\section{\ourdataset Generation}

\begin{table}[!htp]
\centering
\begin{tabular}{ll}
\toprule[1.5pt]
\textbf{Code} & \textbf{Genre} \\
\midrule
\textsf{LF} & Literature \& Fiction \\
\textsf{HB} & History \& Biographies \\
\textsf{BLP} & Business, Law \& Politics \\
\textsf{ST} & Science \& Technology \\
\textsf{ACM} & Art, Culture \& Media \\
\textsf{SSP} & Social Sciences \& Philosophy \\
\textsf{JR} & Journals \& Reports \\
\textsf{LHH} & Lifestyle, Health \& Hobbies \\
\bottomrule[1.5pt]
\end{tabular}
\caption{Different genres used in this work, sourced from Project Gutenberg \citep{momen-etal-2025-filling}.}
\label{tab:genres-info}
\end{table}

\subsection{\Doc Generation Prompts}
\label{sec:novel-gen-prompts}

As described in \refsec{sec:novel-gen}, our \doc generation involves first generating the \doc outline (see Prompt~\ref{prompt:outline}), followed by incremental generation conditioned on the current running summary, outline, and last segment (see Prompts~\ref{prompt:start-novel-segment}~and~{\ref{prompt:continue-novel-segment}}). For summarisation, we adapt the iterative summarisation prompts (see Prompts~\ref{prompt:summarise}~and~\ref{prompt:compress-summary}) from \citet{chang2024booookscore}.

\subsection{Quantitative Analysis}
\label{sec:quantiative-analysis-dataset}

In addition to the standard language understanding metric, perplexity, we evaluate several text diversity metrics as mentioned in \reftab{tab:stats-quantative-eval-dataset}. Considering we are operating in a reference-free setup, we focus on two prominent types of text diversity metrics, similarity between text pairs from the same model, and the token/type ratio \citep{shaib-etal-2025-standardizing}:
\begin{itemize}
    \item \textbf{\selfBleu}: \selfBleu (S-B) measures similarity of each generated text against the remaining texts using BLEU \citep{zhu2018texygen}. It is a homogenisation score widely used to evaluate text diversity in writing tasks \citep{padmakumar2024does}.
    \item \textbf{$n$-Gram Diversity Score}:  $n$-Gram Diversity Score (NGD) \citep{meister2023locally} calculates the ratio of unique $n$-grams to all $n$-grams in the dataset. In our analysis, we consider $n$-grams of length four.
    \item \textbf{Self-repetition}: Self-repetition (S-R) \citep{salkar-etal-2022-self} measures the extent to which n-grams (up to 4-grams) are repeated across different outputs of the same LLM.
    \item \textbf{Compression Ratio}:  Compression Ratio (CR) \citep{shaib-etal-2025-standardizing} defined as the ratio of all LLM-generated texts original file size and to the compressed file size (using gzip), this metric captures redundancy in a manner distinct from $n$-gram-based measures.
    \item \textbf{Perplexity}: Perplexity (PPL) \citep{jm3} is computed as the average negative log-likelihood of the texts using the \texttt{GPT-2-Large} language model as a standard measure of language understanding.
\end{itemize}

\begin{table}[!htp]
    \centering
    \resizebox{0.99\linewidth}{!}{%
    \begin{tabular}{lSSSS[table-format=2.1]}
        \toprule[1.5pt]
        \multirow{2}{*}{\textbf{LLM}} & \multicolumn{4}{c}{\textbf{\Doc Generation}} \\
        \cmidrule(lr){2-5}
            & \multicolumn{1}{c}{Steps} & \multicolumn{1}{c}{Outline} & \multicolumn{1}{c}{Segment} & \multicolumn{1}{c}{Total} \\
        \midrule
        \deepseek       & 39.90 & 0.6K  & 0.8K  & 33.4K \\
        \glm            & 14.12 & 1.8K  & 2.0K  & 28.4K \\
        \kimi           & 15.65 & 2.4K  & 2.4K  & 37.1K \\
        \qwen           & 38.47 & 2.0K  & 1.7K  & 67.1K \\
        \qwenMax        & 27.53 & 1.4K  & 1.5K  & 41.2K \\
        \claude         & 11.96 & 1.7K  & 3.9K  & 47.1K \\
        \geminiFlash    & 8.87  & 2.7K  & 2.3K  & 20.5K \\
        \geminiPro      & 45.85 & 2.2K  & 1.9K  & 88.1K \\
        \gptFour        & 34.00 & 1.2K  & 1.9K  & 66.1K \\
        \gptFive        & 15.92 & 4.1K  & 4.4K  & 69.3K \\
        \bottomrule[1.5pt]
    \end{tabular}
    }
    \caption{\Doc generation statistics across different LLMs. All reported numbers (other than steps) are in words.}
    \label{tab:generation-stat-dataset}
\end{table}

\subsection{Qualitative Analysis}
\label{sec:qualitative-analysis-dataset}

\begin{table*}[htp]
\centering
\resizebox{0.99\textwidth}{!}{
\begin{tabular}{ll SSSSS SSSSS}
\toprule[1.5pt]
\multirow{2}{*}{\textbf{LLM}} & \multirow{2}{*}{\textbf{Genre}} & 
\multicolumn{5}{c}{\textbf{Passage-level}} &
\multicolumn{5}{c}{\textbf{Overall}} \\
\cmidrule(lr){3-7} \cmidrule(lr){8-12}
 & & \multicolumn{1}{c}{Rel.} & \multicolumn{1}{c}{Eng.} & \multicolumn{1}{c}{Coh.} & \multicolumn{1}{c}{Flu.} & \multicolumn{1}{c}{Div.} & \multicolumn{1}{c}{Coh.} & \multicolumn{1}{c}{Emp.} & \multicolumn{1}{c}{Sur.} & \multicolumn{1}{c}{Eng.} & \multicolumn{1}{c}{Comp.} \\
\midrule
\deepseek   & \textsf{SSP} & 3.2 & 2.6 & 3.4 & 3.4 & 4.0 & 3.0 & 3.5 & 4.0 & 3.0 & 3.0 \\
\glm        & \textsf{HB,SSP,JR} & 4.9 & 3.7 & 4.3 & 4.4 & 4.0 & 4.0 & 4.0 & 3.0 & 4.0 & 3.0 \\
\kimi       & \textsf{LF} & 4.8 & 4.4 & 4.2 & 4.2 & 4.4 & 4.0 & 5.0 & 3.0 & 5.0 & 4.0 \\
\qwen       & \textsf{SSP,LF} & 4.6 & 3.7 & 4.7 & 4.8 & 4.7 & 4.0 & 3.0 & 3.5 & 3.5 & 4.0 \\
\qwenMax    & \textsf{LHH,ST} & 3.7 & 3.0 & 4.4 & 4.0 & 3.7 & 3.0 & 3.5 & 4.0 & 2.5 & 2.5 \\
\claude     & \textsf{LF} & 4.0 & 3.9 & 4.2 & 4.7 & 4.5 & 4.0 & 4.0 & 3.5 & 3.5 & 4.0 \\
\geminiFlash& \textsf{LHH,LF} & 4.6 & 3.8 & 4.2 & 4.4 & 4.6 & 4.0 & 4.0 & 3.0 & 4.0 & 2.0 \\
\geminiPro  & \textsf{LHH} & 3.5 & 3.2 & 4.0 & 3.7 & 4.7 & 3.5 & 4.0 & 4.0 & 3.0 & 3.0 \\
\gptFour    & \textsf{HB} & 4.3 & 3.7 & 4.3 & 4.2 & 4.4 & 4.0 & 3.0 & 3.0 & 3.0 & 3.5 \\
\gptFive    & \textsf{ACM} & 4.4 & 3.6 & 5.0 & 4.8 & 4.8 & 4.0 & 4.0 & 3.0 & 3.0 & 4.0 \\
\midrule
\multicolumn{2}{c}{\textbf{Overall}} 
           & 4.2
           & 3.6
           & 4.3
           & 4.3
           & 4.4
           & 3.8
           & 3.8
           & 3.4
           & 3.5
           & 3.3 \\
\bottomrule[1.5pt]
\end{tabular}}
\caption{Human evaluation scores across passage-level and overall narrative quality dimensions for different LLMs.
Rel. - Relevance, Eng. - Engagement, Coh. - Coherence, Flu. - Fluency, Div. - Diversity,  Emp. - Empathy, Sur. - Surprise, and Comp. - Complexity. 1 is the lowest and 5 is the highest score.
}
\label{tab:human_eval_llms}
\end{table*}

Following \citet{chhun2022human,wang2025towards}, we evaluate LLM-generated books across several dimensions capturing meso-, micro-, and macro-level properties.
\begin{itemize}
    \item \textbf{Relevance (Rel.)}: how well the text aligns with or follows the given instruction. Ensures that the generated text is appropriate and follows the given book outline.
    \item \textbf{Engagement (Eng.)}: measures the degree to which the reader is engaged with the book.
    \item \textbf{Coherence (Coh.)}: measures whether the story makes sense. Ensures the generated text flows logically from beginning to end, with consistent narrative elements and clear progression.
    \item \textbf{Fluency (Flu.)}: measures the smoothness and natural flow of the language.
    \item \textbf{Diversity (Div.)}: measures variety in linguistic expression, including rich vocabulary and varied sentence structures.
    \item \textbf{Empathy (Emp.)}: measures how well the model captures the characters' emotions, regardless of whether the reader agrees with them.
    \item \textbf{Surprise (Sur.)}: measures how surprising the ending of the book is.
    \item \textbf{Complexity (Comp.)}: measures how elaborate the book is.
\end{itemize}

\paragraph{Annotators.}
The annotators were volunteers from our university representing mixed demographics. All annotators were fluent in English, and we evaluated $10$ books, one for each LLM. Annotators were provided with clear instructions for the assigned task. No compensation beyond voluntary participation was offered, and all annotators were fully informed about the nature of the study and their role. 

\paragraph{Annotation Task.}
Given that our generated books are long, we randomly sample five passages capturing the flow of each book. Each passage comprises five questions, followed by five questions at the end of the book, totalling 30 questions per \doc (results in \reftab{tab:human_eval_llms}). Each question is answered on a 5-point Likert scale.  We acknowledge potential confounding factors (such as annotator, genre, and model). The human evaluation is intended primarily to provide confidence that the generated text is reasonable.

\paragraph{Results.}
\reftab{tab:human_eval_llms} presents the average annotator scores across different quality dimensions for both at the passage level and overall for each \doc. We note strong scores ($>4$; out of 5) for relevance, coherence, fluency, and diversity, complementing the text quality and diversity analysis from \reftab{tab:stats-quantative-eval-dataset}. For other subjective dimensions (engagement, empathy, surprise, and complexity), the scores are also generally positive ($>3.3$). This indicates that the texts in \ourdataset exhibit {good quality} qualitatively.
Annotators' agreement (Krippendorff's $\alpha=0.17$) remains low across evaluation tasks.
We attribute this to the highly subjective nature of the task and the length of the generations (49K words on average), which can make it difficult for annotators to follow the narrative consistently.  However, 70\% of ratings were within $\pm 1$ Likert point, with a mean difference of 1.05, indicating that most disagreements were relatively minor.

\begin{table}[!htp]
    \centering
    \begin{tabular}{lSSS}
        \toprule[1.5pt]
        \multirow{2}{*}{\textbf{LLM}} & \multicolumn{2}{c}{API Cost (\$/$1$M)}  & {\textbf{Total}} \\
        \cmidrule(lr){2-3}
            & \multicolumn{1}{c}{Input}  & \multicolumn{1}{c}{Output} & {(in \$)} \\
        \midrule
        \deepseek    & 0.2 & 0.5  & 10.6 \\
        \glm         & 0.5 & 1.8  & 10.9 \\
        \kimi        & 0.5 & 2.0  & 22.1 \\
        \qwen        & 0.1 & 0.1  & 9.0  \\
        \qwenMax     & 1.2 & 6.0  & 60.2 \\
        \claude      & 3.0 & 10.0 & 85.9 \\
        \geminiFlash & 0.3 & 2.5  & 6.6  \\
        \geminiPro   & 1.3 & 10.0 & 155.0 \\
        \gptFour     & 2.0 & 8.0  & 81.4 \\
        \gptFive     & 1.3 & 10.0 & 88.7 \\
         \midrule
         \multicolumn{3}{c}{Total} & 530.4 \\
         \bottomrule[1.5pt]
    \end{tabular}
    \caption{\ourdataset generation cost breakdown. `Input' and `Output' denote API cost per $1$M tokens. `Total' is the cost of generating all books for a given LLM (as per the statistics in \reftab{tab:stats-quantative-eval-dataset}). \texttt{OpenRouter} API costs are as of September 2025. }
    \label{tab:dataset-API-cost-breakdown}
\end{table}

\subsection{Cost Analysis}
\refapptab{tab:dataset-API-cost-breakdown} lists generation costs per LLM. In total, generating \ourdataset cost approximately \$$530$.

\begin{figure}[!tp]
    \centering
    \includegraphics[width=0.99\linewidth]{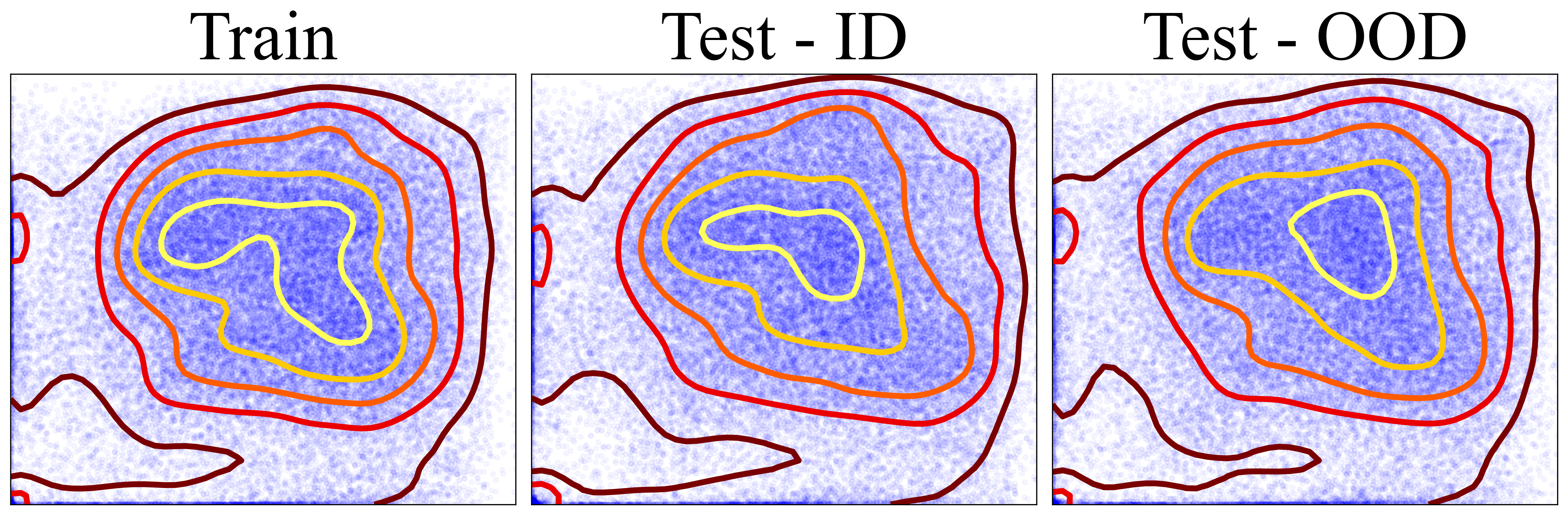}
    \caption{An example where \ourmethod captures similar \gptFour fingerprints across domains, demonstrating generalisation. Left: a training reference fingerprint; Middle: a test sample from seen domain (\textsf{HB}); Right: a test sample from an unseen domain (\textsf{SSP,LHH}). 
    }
    \label{fig:example-fingperprints}
\end{figure}

\section{\ourmethod Fingerprint}

\subsection{Similarity Metrics}
\label{sec:sim-metrics}
We formulate the two similarity metrics used to compare two fingerprints, $f_A$ and $f_B$, below:

\paragraph{JS Distance.}
\[
\operatorname{JS} = 
\sqrt{
\frac12\, D_{\mathrm{KL}}(f_A \,\|\, M)
+
\frac12\, D_{\mathrm{KL}}(f_B \,\|\, M)
}
\]
with
\[
M = \frac{f_A + f_B}{2}, \quad
D_{\mathrm{KL}}(p \,\|\, q) = \sum_k p_k \log_2 \frac{p_k}{q_k}.
\]
\paragraph{Norm-Mean.}
Compute gradients of fingerprint $f$:
\[
\begin{aligned}
f_x(i,j) &\approx \frac{f(i,j+1) - f(i,j-1)}{2\,\mathrm{d}x},\\
f_y(i,j) &\approx \frac{f(i+1,j) - f(i-1,j)}{2\,\mathrm{d}y}.
\end{aligned}
\]
Using gradients, we compute surface normals:
\[
\begin{aligned}
\hat{\mathbf{n}}(i,j) = \frac{(-f_x(i,j), -f_y(i,j), 1)}{\sqrt{f_x(i,j)^2 + f_y(i,j)^2 + 1}}.
\end{aligned}
\]
Mean angular difference between fingerprints $f_A$ and $f_B$:
\[
\operatorname{norm-mean} = \frac{1}{n^2} \sum_{i,j=1}^n 
\arccos\bigl(\hat{\mathbf{n}}_A(i,j) \cdot \hat{\mathbf{n}}_B(i,j)\bigr),
\]
where $\hat{\mathbf{n}}_A$ and $\hat{\mathbf{n}}_B$ are computed from $f_A$ and $f_B$.

\section{Experimental Details (cont.)}
\label{sec:app-exp-details}

We use \texttt{OpenRouter}\footnote{\url{https://openrouter.ai/}} for all LLM API calls used for \ourdataset generation, with \texttt{temperature} $=1.0$, \texttt{top-P} $=1.0$, \texttt{top-K} $=0$, and \texttt{maximum\_output\_length} $=8192$ tokens. 
We provide a model card for all LLMs used in this work in \reftab{tab:model-card}.
All experiments were conducted using a single \texttt{A100 GPU} with \texttt{CUDA 12.4} and \texttt{PyTorch 2.10.0}. 

For \bert and \topformer, as they do not process long-form documents by default, we split each text into $512$-token chunks and average the prediction scores to compute the final prediction, following \citet{tyo2023valla}. For \detective, since it is clustering-based, we compute the percentage of top-$K$ results across text chunks. As noted earlier, all hyperparameter tuning was performed on the development set, with $5$ epochs and $20$ epochs of fine-tuning for both \bert and \topformer and \detective, respectively. In Rank-based fingerprint, we set $\alpha = 1.5$ and use $100$K samples for power-law approximation (more in \refappalg{algo:rank-transition-fingerprint}). For all \ourmethod experiments, we use \gpt as evaluator language model using context of $1024$ and $50$ grid size (or cluster for ranks) unless stated otherwise. 

\begin{table*}[!htp]
    \centering
    \resizebox{0.99\textwidth}{!}{
    \begin{tabular}{llS[table-format=3.1]S[table-format=3.1]ll}
    \toprule[1.5pt]
    \textbf{Model Name} & \textbf{\texttt{OpenRouter} Checkpoint} & \textbf{Context} & \textbf{Max. Out.} & \textbf{Type} & \textbf{Reference} \\ 
    \midrule
    \deepseek & {\footnotesize\texttt{deepseek/deepseek-chat-v3-0324}} & {163.8K} & {163.8K} & {\scalebox{0.9}{\faLockOpen}\ Open} & {\footnotesize \citet{deepseek-v3}} \\
    \glm      & {\footnotesize\texttt{z-ai/glm-4.6}} & {204.8K} & {204.8K} & {\scalebox{0.9}{\faLockOpen}\ Open} & {\footnotesize \citet{glm-4.6}} \\
    \kimi     & {\footnotesize\texttt{moonshotai/kimi-k2-thinking}} & {131.1K} & {131.1K} & {\scalebox{0.9}{\faLockOpen}\ Open} & {\footnotesize \citet{kimi-k2}} \\
    \qwen     & {\footnotesize\texttt{qwen/qwen3-235b-a22b-2507}} & {262.1K} & {262.1K} & {\scalebox{0.9}{\faLockOpen}\ Open} & {\footnotesize \citet{qwen-3}} \\
    \midrule
    \qwenMax  & {\footnotesize\texttt{qwen/qwen3-max}} & {262.1K} & {32.8K} & {\scalebox{0.9}{\faLock}\ Closed} & {\footnotesize \citet{qwen3-max}} \\
    \claude   & {\footnotesize\texttt{anthropic/claude-sonnet-4.5}} & {1M} & {64K} & {\scalebox{0.9}{\faLock}\ Closed} & {\footnotesize \citet{claude-sonnet-4.5}} \\
    \geminiFlash & {\footnotesize\texttt{google/gemini-2.5-flash}} & {1.05M} & {65.5K} & {\scalebox{0.9}{\faLock}\ Closed} & {\footnotesize \citet{gemini-2.5}} \\
    \geminiPro   & {\footnotesize\texttt{google/gemini-2.5-pro}} & {1.05M} & {65.5K} & {\scalebox{0.9}{\faLock}\ Closed} & {\footnotesize \citet{gemini-2.5}} \\
    \gptFour  & {\footnotesize\texttt{openai/gpt-4.1}} & {1.05M} & {32.8K} & {\scalebox{0.9}{\faLock}\ Closed} & {\footnotesize \citet{gpt-4}} \\
    \gptFive  & {\footnotesize\texttt{openai/gpt-5.1}} & {400K} & {128K} & {\scalebox{0.9}{\faLock}\ Closed} & {\footnotesize \citet{gpt-5}} \\
    \bottomrule[1.5pt]
    \end{tabular}
    }
    \caption{Details for all the LLMs used in \ourdataset. `Context' is the total input context supported and `Max. Out.' is the maximum number of output tokens supported by the LLM. `Type' indicates whether model weights are public ({\scalebox{0.9}{\faLockOpen}\ Open}) or if it is proprietary LLM ({\scalebox{0.9}{\faLock}\ Closed}).
    }
    \label{tab:model-card}
\end{table*}

\section{Baselines (cont.)}
\label{sec:baselines}

\paragraph{Rank.} We compute the average token rank for a given text using the \texttt{GPT-2-Medium} model.

\paragraph{Entropy.} Similarly, we compute the average token entropy for a given text using the evaluator model, \texttt{GPT-2-Medium}.

\paragraph{GLTR.} GLTR \citep{gehrmann2019gltr} computes the probability of a token falling into four rank ranges (or buckets). Compared with a single average-based metric, it is more representative. 

For all three above metric-based techniques, we adapt them to the AA task following \citet{la2025openturingbench} by attaching a logistic regression classifier on top of these metric values.

\paragraph{\ngram.} This is a traditional but effective technique that ensembles three $n$-gram models: character-based, part-of-speech-based, and summary statistics. Moreover, it can naturally handle long texts, so no chunking is required, unlike the next two methods.

\paragraph{\bert.} A multi-class classification layer is added to the pre-trained BERT model \cite{fabien-etal-2020-bertaa}, which is then fine-tuned on a specialised dataset. However, as a Transformer-based architecture \citep{vaswani2017attention}, it has a bottleneck of processing at most $512$ tokens at a time \citep{devlin_bert_2019}.

\addTwo{
\paragraph{\topformer.} \textbf{Top}ological Trans\textbf{former}-based model (\topformer) \citep{uchendu2024topformer} attaches topological data analysis (TDA) layer on top of pre-trained \textsc{RoBERTa}. The motivation is that TDA is known to capture shape and structure (\ie linguistic structure) enhancing contextual representations of base model, which is fine-tuned similar to \bert.}

\paragraph{\detective.} \addTwo{This is a multi-level contrastive learning–based state-of-the-art method for binary AI detection \citep{guo2024detective}. However, it internally learns representations for each author. Therefore, these embeddings can also be used for the AA task. We also considered more recent \textsc{M-RangeDetector} \citep{jiao-etal-2025-rangedetector}, \addThree{\textsc{PhantomHunter} \citep{shi2025phantomhunterdetectingunseenprivatelytuned}}, and other simpler contrastive learning methods \citep{la2024contrasting,malyalam-contrastive-AA}, but their implementations are not publicly available, and we were unable to obtain sufficient details for reproduction.}

\begin{algorithm}[!htp]
\caption{Rank-based Fingerprint ($f^{\operatorname{rank}}$) Compression}
\begin{algorithmic}[1]
\Require $|\mathcal{V}|$, $R = \langle r_1, \dots, r_m \rangle$, $\alpha$ (power-law exponent), $C$ (clusters)

\State $p(i) \gets \text{power-law approx. with } \alpha, \;\forall i \in \{1\dots|\mathcal{V}|\}$

\State $\text{cum} \gets 0$, $k \gets 1$, $\Delta \gets 1/C$

\For{$i \gets 1$ to $|\mathcal{V}|$}
    \State $\text{cluster}[i] \gets k$
    \State $\text{cum} \gets \text{cum} + p(i)$
    \While{$\text{cum} \geq \Delta$}
        \State $\text{cum} \gets \text{cum} - \Delta$
        \State $k \gets k + 1$
    \EndWhile
\EndFor

\State $K \gets k$ \Comment{number of used clusters}
\State $f^{\operatorname{rank}} \gets \mathbf{0}^{K \times K}$

\For{$i \gets 2$ to $m$}
    \State $f^{\operatorname{rank}}[\text{cluster}[R_{i-1}], \text{cluster}[R_i]] \mathrel{+}= 1$
\EndFor

\State \Return $f^{\operatorname{rank}}$ \Comment{$\mathbb{R}^{K \times K}$}
\end{algorithmic}
\label{algo:rank-transition-fingerprint}
\end{algorithm}
\begin{algorithm}[!htp]
\caption{Entropy-based Fingerprint ($f^{\operatorname{entropy}}$) Smoothing}
\begin{algorithmic}[1]
\Require $|\mathcal{V}|$, $E = \langle e_1,\dots,e_m \rangle$, $G$ (grid size)

\State $P \gets \{(e_{i-1},e_i) \mid i=2,\dots,m\}$
\State $\text{KDE} \gets \text{GaussianKDE}(P)$
\State $x_\text{grid},y_\text{grid} \gets \text{linspace}(0,\log(|\mathcal{V}|),G)$
\State $f^{\operatorname{entropy}} \gets \text{KDE}(\text{meshgrid}(x_\text{grid},y_\text{grid}))$
\State \Return $f^{\operatorname{entropy}}$ \Comment{$\mathbb{R}^{G \times G}$}
\end{algorithmic}
\label{algo:entropy-estimation}
\end{algorithm}

\begin{table}[!htp]
\centering
\resizebox{0.99\columnwidth}{!}{
\begin{tabular}{l
                l
                l@{}
                S[table-format=2.0]}
\toprule[1.5pt]
\textbf{Model} & \textbf{Split} & \textbf{Genres} & \textbf{\# Books} \\
\midrule
\multirow{3}{*}{\deepseek}
  & ID & \textsf{LF} & 33 \\
  \cmidrule(lr){2-4}
  & \oodOne & \textsf{LHH, HB} & 2 \\
  & & \textsf{BLP, HB} & 2 \\
  & & \textsf{SSP} & 5 \\
\midrule
\multirow{3}{*}{\glm}
  & ID & \textsf{HB, SSP, JR} & 3 \\
  & & \textsf{HB, JR} & 2 \\
  & & \textsf{SSP, LF} & 15 \\
  & & \textsf{LF} & 6 \\
  \cmidrule(lr){2-4}
  & \oodOne & \textsf{ST} & 6 \\
\midrule
\multirow{2}{*}{\kimi}
  & ID & \textsf{LF} & 30 \\
  \cmidrule(lr){2-4}
  & \oodOne & \textsf{HB} & 5 \\
  & & \textsf{ST} & 2 \\
\midrule
\multirow{4}{*}{\qwen}
  & ID & \textsf{LF} & 11 \\
  & & \textsf{SSP, LF} & 19 \\
  \cmidrule(lr){2-4}
  & \oodOne & \textsf{BLP, LHH, HB} & 1 \\
  & & \textsf{BLP, HB} & 3 \\
  & & \textsf{HB} & 4 \\
\midrule
\multirow{4}{*}{\qwenMax}
  & ID & \textsf{LHH, ST} & 11 \\
  & & \textsf{SSP} & 11 \\
  & & \textsf{SSP, ACM} & 6 \\
  \cmidrule(lr){2-4}
  & \oodOne & \textsf{LF} & 6 \\
\midrule
\multirow{3}{*}{\claude}
  & ID & \textsf{HB, LF} & 11 \\
  & & \textsf{LF} & 8 \\
  \cmidrule(lr){2-4}
  & \oodOne & \textsf{BLP, SSP} & 5 \\
\midrule
\multirow{3}{*}{\geminiFlash}
  & ID & \textsf{HB, ACM} & 7 \\
  & & \textsf{SSP} & 17 \\
  \cmidrule(lr){2-4}
  & \oodOne & \textsf{LHH, LF} & 6 \\
\midrule
\multirow{3}{*}{\geminiPro}
  & ID & \textsf{LHH, HB} & 5 \\
  & & \textsf{HB} & 6 \\
  & & \textsf{LHH} & 11 \\
  \cmidrule(lr){2-4}
  & \oodOne & \textsf{LF} & 5 \\
\midrule
\multirow{2}{*}{\gptFour}
  & ID & \textsf{HB} & 29 \\
  \cmidrule(lr){2-4}
  & \oodOne & \textsf{LHH, SSP} & 6 \\
\midrule
\multirow{5}{*}{\gptFive}
  & ID & \textsf{SSP, ACM} & 7 \\
  & & \textsf{ACM} & 7 \\
  & & \textsf{SSP} & 7 \\
  \cmidrule(lr){2-4}
  & \oodOne & \textsf{ST, LF} & 2 \\
  & & \textsf{LHH, ST} & 3 \\
\bottomrule[1.5pt]
\end{tabular}}
\caption{Detailed genre splits (ID \versus \oodOne) for different LLMs in \ourdataset, complementing \reftab{tab:stats-quantative-eval-dataset}. These genre splits replicate human book distribution from Project Gutenberg.}
\label{tab:detailed-dataset-statistics}
\end{table}

\section{Further Ablation Studies}
\label{sec:app-ablation-studies}

\paragraph{Top-$k$ Accuracy AA Performance.}
\add{
Following prior work in authorship attribution \citep{huang2025authorship,uchendu-etal-2020-authorship}, we additionally report top-$k$ accuracy ($k=1,2,3,5$) as a complementary evaluation metric. Since there are only ten authors in our case, top-$k$ accuracy saturates quickly as seen in \reftab{tab:all_authors-top-k-results}. We nevertheless include it for completeness, and observe that it is consistent with the main findings reported using \fscore in \reftab{tab:all_authors-main-results}.}

\paragraph{\oodTwo Additional Analysis.}
\addThree{
To further strengthen our \oodTwo evaluation, we created an additional balanced test set containing both unknown and known (randomly sampled from same split) generators, resembling a mixed deployment scenario, reported in \reftab{tab:mix-set-ood-A-eval}.}
\addThree{We also conducted error analysis, evaluating per-LLM \fscore \oodTwo performance in \reftab{tab:per-LLM-OOD-AA-perf} to check whether performance is dominated by a few LLMs. It shows that the macro-average is usually not driven by a single or two models for TRACE and is more balanced compared to other methods. It further highlights substantial variation in difficulty across LLMs (\eg \geminiPro, the \texttt{Qwen} family are generally more challenging). Overall, these results further strengthen and complement the main \oodTwo results from \reftab{tab:main-results}.}

\begin{table}[!htp]
    \centering
    \begin{tabular}{lcc}
        \toprule[1.5pt]
        \multirow{2}{*}{\textbf{Method}} & \multicolumn{2}{c}{\textbf{\oodTwo}}  \\
        \cmidrule(lr){2-3}
         & \textbf{AUROC} & \textbf{AUPR} \\
        \midrule
        \textbf{\rank} & $0.49$ & $0.62$ \\
        \textbf{\entropy} & $0.58$ & $0.66$ \\
        \textbf{\gltr} & $0.51$ & $0.63$ \\
        \midrule
        \textbf{\ngram} & \underline{$0.90$} & \underline{$0.89$} \\
        \textbf{\bert} & $0.83$ & $0.86$ \\
        \textbf{\topformer} & $0.80$ & $0.84$ \\
        \textbf{\detective} & $0.81$ & $0.84$ \\
        \midrule
        \textbf{\ourmethod (Ours)} \\
        \ \ Ranks-JS & $0.72$ & $0.74$ \\
        \ \ Entropy-JS & \bm{$0.92$} & \bm{$0.93$} \\
        \ \ Entropy-Norm & \bm{$0.92$} & \bm{$0.93$} \\
        \bottomrule[1.5pt]
    \end{tabular}
    \caption{\addThree{\oodTwo performance on balanced test set containing both unknown and known author samples. Reported numbers are average threshold-independent metrics (AUROC and AUPR) over three random splits. Entropy-based \ourmethod is remains performant as seen in the main results.}}
    \label{tab:mix-set-ood-A-eval}
\end{table}

\begin{table*}[!htp]
    \centering
    \resizebox{0.95\textwidth}{!}{
    \begin{tabular}{l ccccccc | c}
        \toprule[1.5pt]
        \multirow{2}{*}{\textbf{Method}} & \multirow{2}{*}{\textbf{\ngram}} & \multirow{2}{*}{\textbf{\bert}} & \multirow{2}{*}{\textbf{\textsc{DeTeCt.}}} & \multirow{2}{*}{\textbf{\textsc{TopForm.}}} & \multicolumn{3}{c}{\textbf{\ourmethod (Ours)}} & \multirow{2}{*}{\textbf{Avg.}}  \\
        \cmidrule(lr){6-8}
         & & & & & {Ranks-JS} & {Entropy-JS} & {Entropy-Norm} & \\
         \midrule
            \deepseek & $0.31$ & $0.22$ & $0.26$ & $0.00$ & $0.95$ & $0.99$ & $0.99$ & $0.53$ \\
            \glm & $0.28$ & $0.14$ & $0.46$ & $0.25$ & $0.32$ & $0.31$ & $0.28$ & $0.29$ \\
            \kimi & $0.33$ & $0.28$ & $0.80$ & $0.84$ & $0.40$ & $0.32$ & $0.31$ & $0.47$ \\
            \qwen & $0.33$ & $0.00$ & $0.22$ & $0.22$ & $0.00$ & $0.22$ & $0.44$ & $0.20$ \\
            \qwenMax & $0.54$ & $0.36$ & $0.32$ & $0.27$ & $0.00$ & $0.17$ & $0.05$ & $0.24$ \\
            \claude & $0.22$ & $0.07$ & $0.58$ & $0.59$ & $0.19$ & $0.98$ & $0.96$ & $0.51$ \\
            \geminiFlash & $0.44$ & $0.00$ & $0.56$ & $0.00$ & $0.00$ & $0.78$ & $0.67$ & $0.35$ \\
            \geminiPro & $0.17$ & $0.21$ & $0.21$ & $0.17$ & $0.00$ & $0.33$ & $0.12$ & $0.17$ \\
            \gptFour & $0.27$ & $0.22$ & $0.37$ & $0.43$ & $0.05$ & $0.74$ & $0.23$ & $0.33$ \\
            \gptFive & $0.29$ & $0.17$ & $0.78$ & $1.00$ & $0.00$ & $0.51$ & $0.34$ & $0.44$ \\
         \bottomrule[1.5pt]
    \end{tabular}
    }
    \caption{\oodTwo per-LLM \fscore scores (averaged over the three splits). `Avg.' column  is row-wise average, highlighting some LLMs are particularly more difficult to differentiate. }
    \label{tab:per-LLM-OOD-AA-perf}
\end{table*}

\paragraph{Impact of Fingerprint Size and Order.}
\begin{figure}[!htp]
    \centering
    \includegraphics[width=0.95\linewidth]{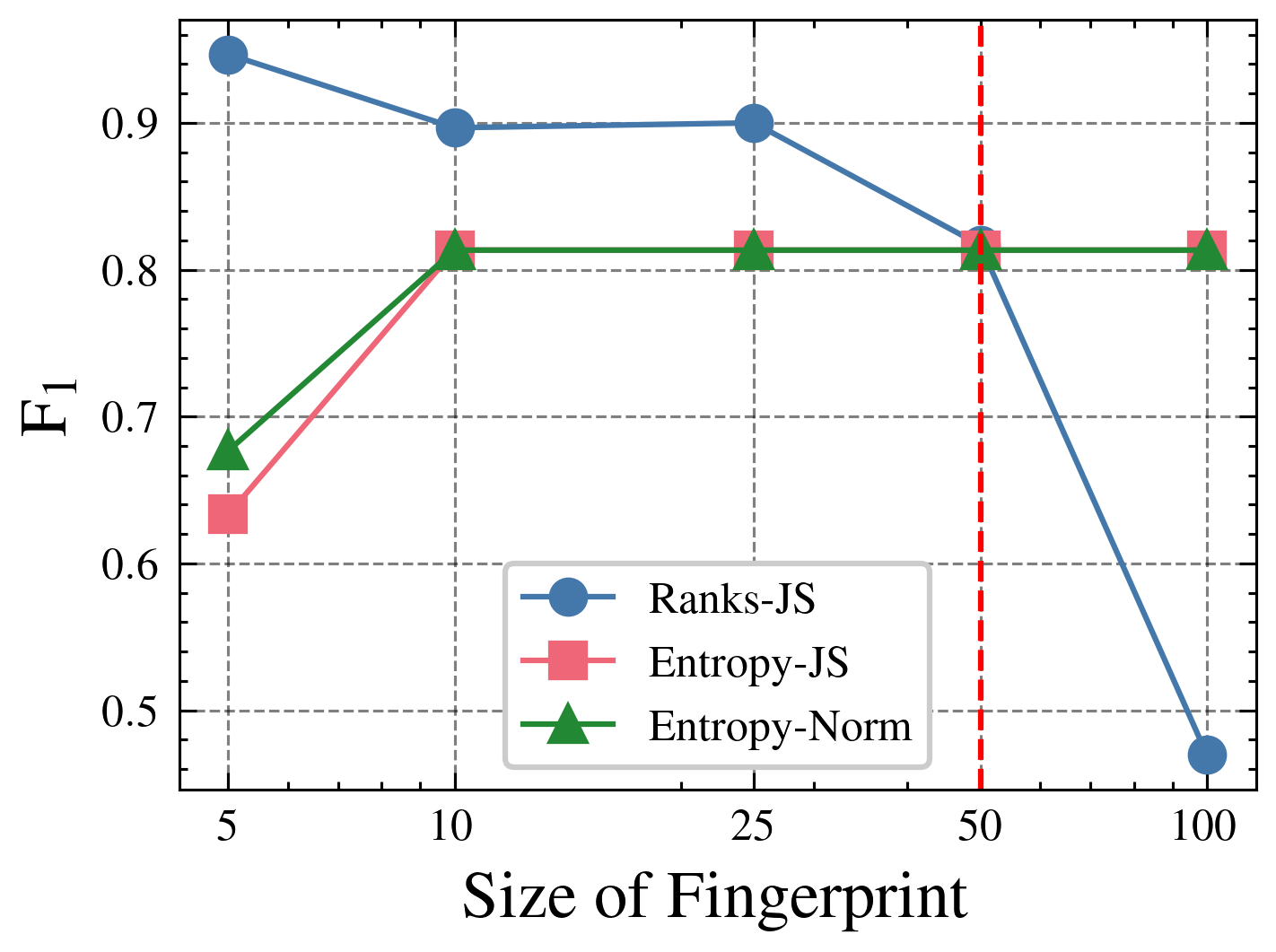}
    \caption{The impact of using different fingerprint sizes \add{($K$ clusters for Rank-based and $G$ grid-size for Entropy-based)} for \ourmethod. The \textcolor{red}{red} dashed line denotes the selected configuration.}
    \label{fig:impact-fingperprint-size}
\end{figure}

\begin{figure}[!htp]
    \centering
    \begin{subfigure}{0.95\columnwidth}
        \includegraphics[width=\linewidth]{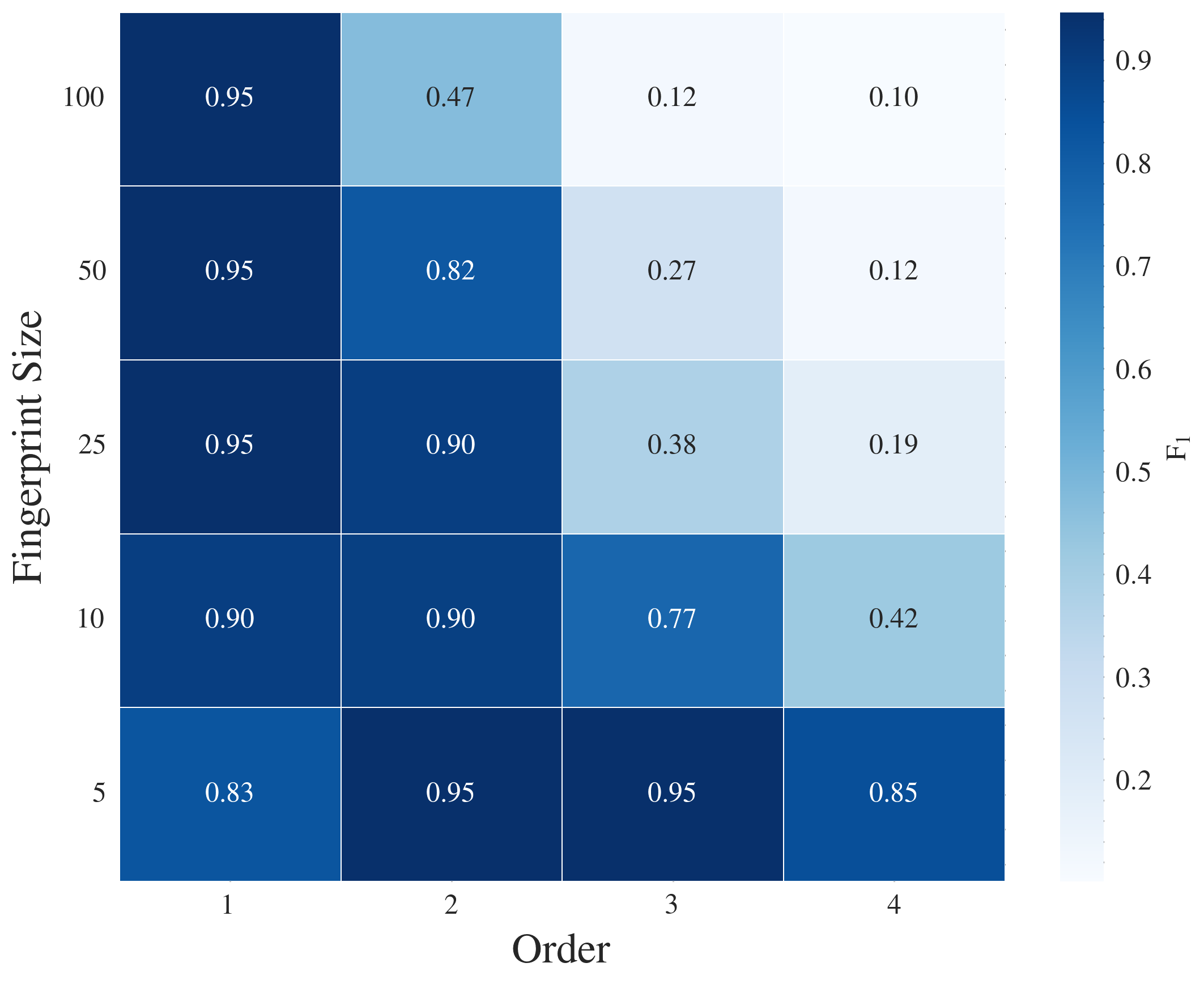}
        \caption{Rank-based}
    \end{subfigure}
    \begin{subfigure}{0.9\columnwidth}
        \includegraphics[width=\linewidth]{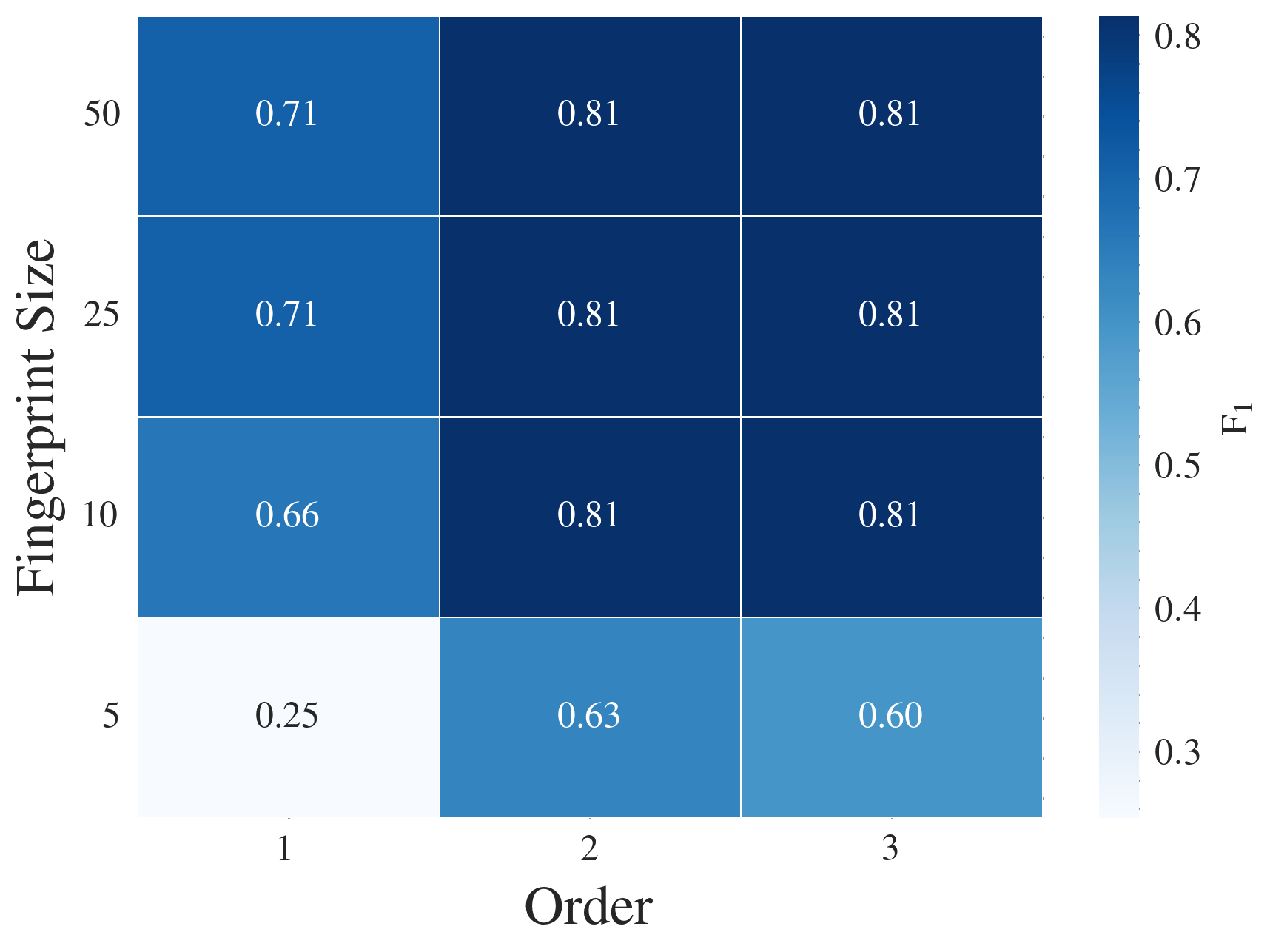}
        \caption{Entropy-Based 
        }
    \end{subfigure}
    \caption{The impact of different orders used in \ourmethod fingerprints. 
    \label{fig:fingperprint-order}
    }
\end{figure}

We investigate different fingerprint sizes for Rank-based and Entropy-based \ourmethod variants. As the cluster-based compression is applied over a large range of ranks, the size and order of the fingerprint can be important factors. From \reffig{fig:impact-fingperprint-size}, we observe that lower fingerprint sizes (\ie grid size $G$ from \refsec{sec:entropy-based-fingerprint}) perform poorly for the Entropy-based variant, whereas larger sizes (\ie clusters $K$ from \refsec{sec:rank-based-fingerprint}) are less effective for the Rank-based variant. A higher number of clusters when compressing ranks may not sufficiently reduce sparsity. A similar trend appears in \refappfig{fig:fingperprint-order}, where increasing the transition order (\eg from $2^{\text{nd}}$ order ($r_{i-1} \rightarrow r_i$; default) to $3^{\text{rd}}$ order transitions ($r_{i-2} \rightarrow r_{i-1} \rightarrow r_{i}$)) leads to increased sparsity and reduced performance.

\begin{figure}[!htp]
    \centering
    \includegraphics[width=0.95\linewidth]{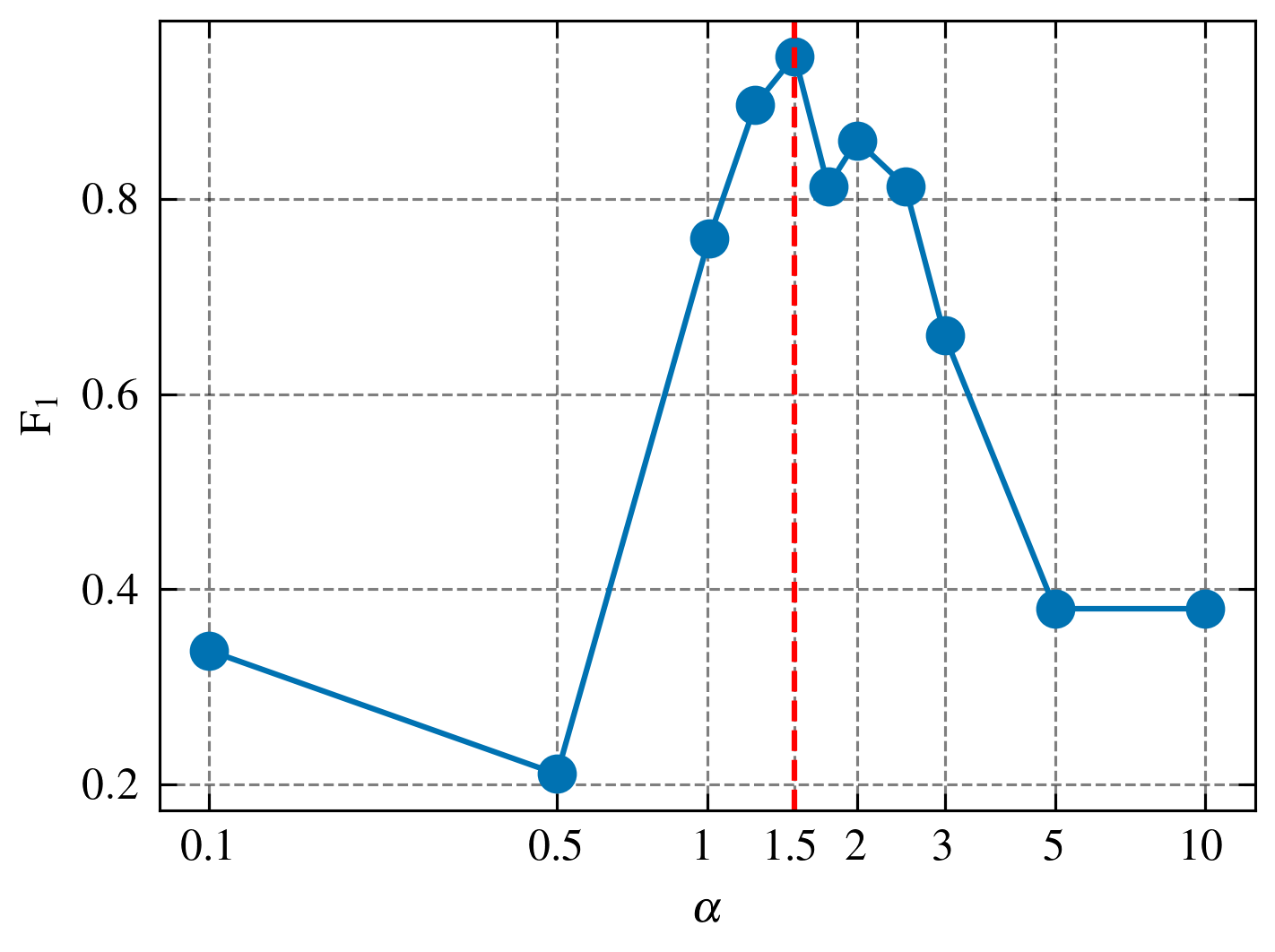}
    \caption{The impact of $\alpha$ in power law approximation for Rank-based \ourmethod. The \textcolor{red}{red} dashed line denotes the chosen $\alpha=1.5$.
    \label{fig:diff-alpha}
    }

\end{figure}
\paragraph{Impact of $\alpha$ in Rank-based Fingerprint.}

Considering the sparsity in Rank-based fingerprints due to the large vocabulary size, we perform compression clustering by grouping ranks into equal-probability bins, as outlined in \refappalg{algo:rank-transition-fingerprint}. We approximate ranks distribution as power-law distribution using the parameter $\alpha$. We investigate the impact of $\alpha$ in \ourmethod. As expected, for $\alpha < 1$ or $\alpha > 5$, performance degrades because the distribution no longer resembles a power law. Fortunately, a range of $\alpha$ values can work, as seen in \reffig{fig:diff-alpha}; we use $\alpha = 1.5$ unless stated otherwise.

\begin{figure}[!htp]
    \centering
    \includegraphics[width=0.95\linewidth]{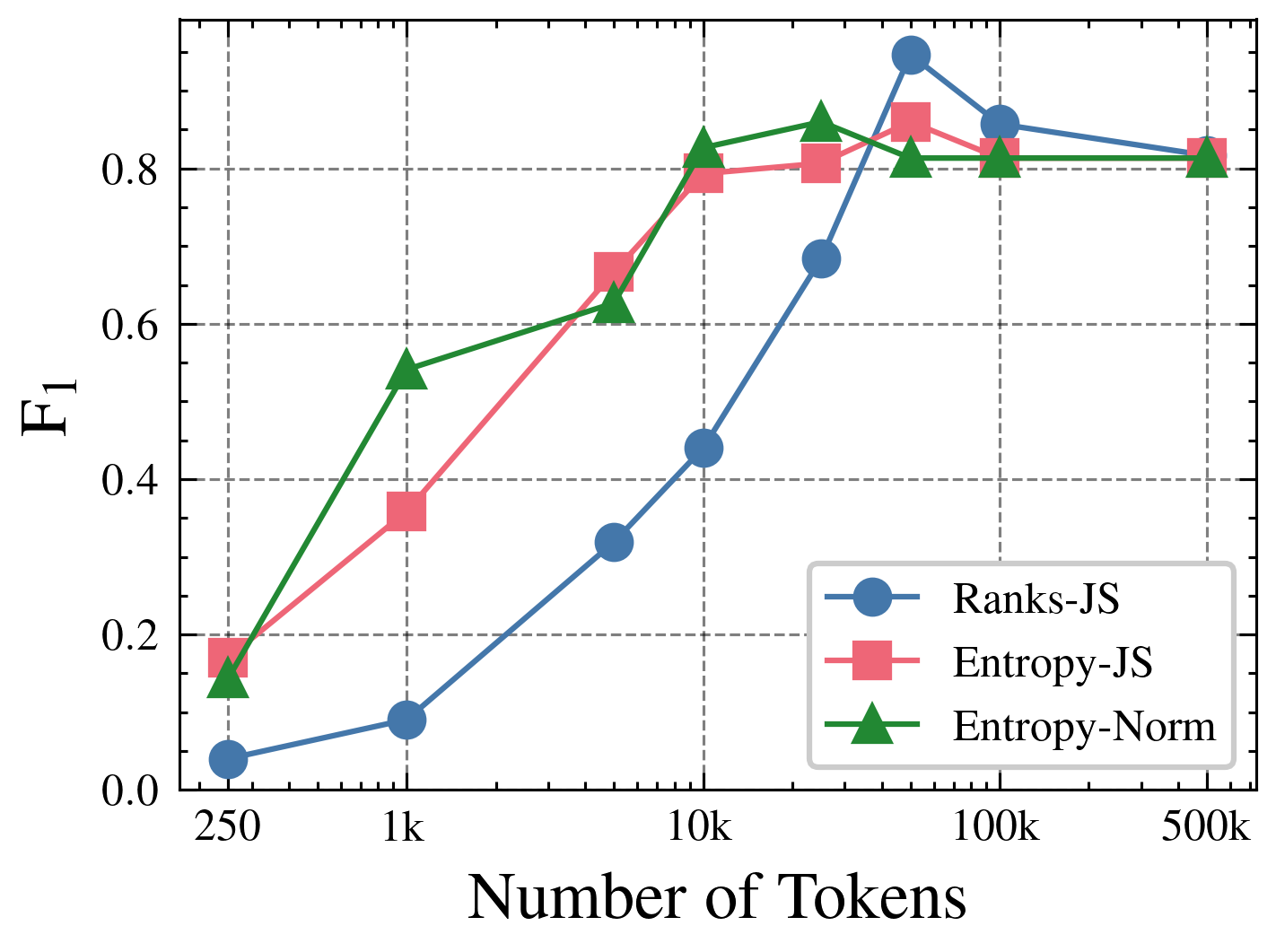}
    \caption{The impact of the number of tokens considered per text in \ourmethod.
    }
    \label{fig:num-tokens}
\end{figure}

\paragraph{Impact of Number of Tokens.}
Although this paper studies a long-form document, we investigate how limiting the number of tokens (per book) impacts the performance of \ourmethod. From \reffig{fig:num-tokens}, we can see that for Entropy-based fingerprints, the performance converges at $10$K tokens, whereas for Rank-based, it converges at $50$K. This suggests \ourmethod works for shorter documents as well; we leave it for future work.

\begin{figure}[!htp]
    \centering
    \includegraphics[width=0.95\linewidth]{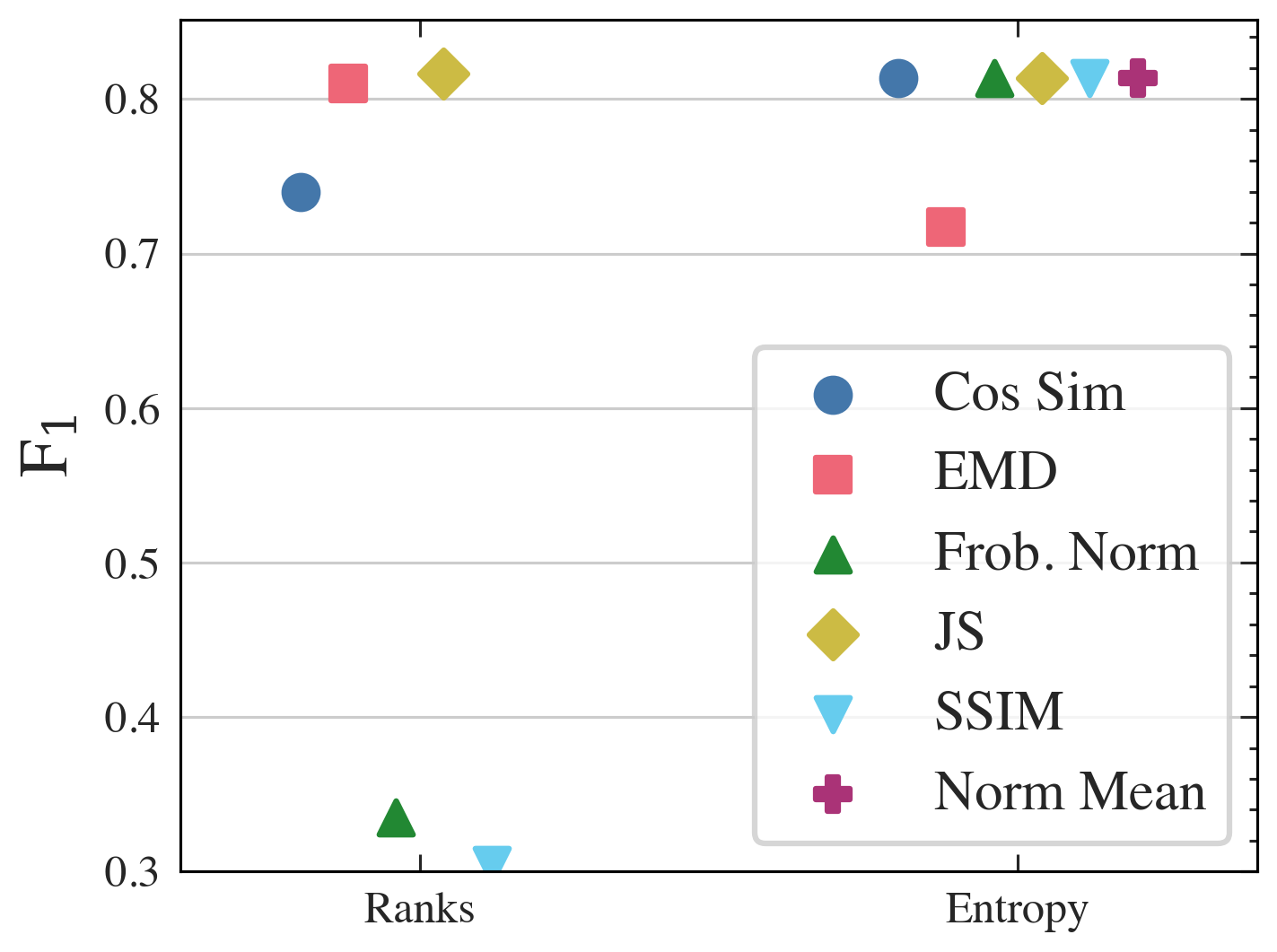}
    \caption{Evaluation of different metrics for comparing \ourmethod fingerprints.}
    \label{fig:scatter-diff-metrics}
\end{figure}

\paragraph{Different Similarity Metrics.} 
We investigate several metrics for comparing two \ourmethod fingerprints in \reffig{fig:scatter-diff-metrics}. We observe that JS distance (along with Wasserstein or Earth-Mover distance) is effective for Rank-based fingerprints. This is expected when these fingerprints are viewed as discrete probability functions. In contrast, for Entropy-based fingerprints, the differences are small, and most metrics perform similarly. Future work could explore information-geometric metrics to further improve discrimination.

\begin{figure*}[!htp]
    \centering
    \begin{subfigure}{0.32\textwidth}
        \includegraphics[width=\textwidth]{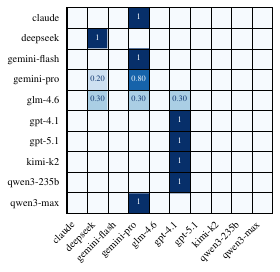}
        \caption{\gltr}
    \end{subfigure}
    \begin{subfigure}{0.32\textwidth}
        \includegraphics[width=\textwidth]{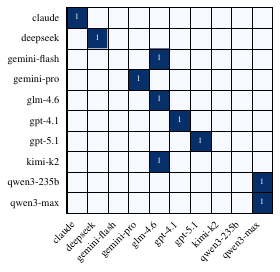}
        \caption{\ngram}
    \end{subfigure}
    \begin{subfigure}{0.32\textwidth}
        \includegraphics[width=\textwidth]{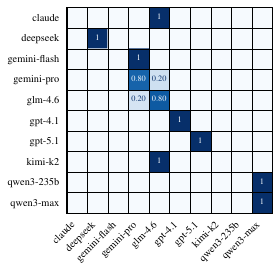}
        \caption{\detective}
    \end{subfigure}
    \begin{subfigure}{0.32\textwidth}
        \includegraphics[width=\textwidth]{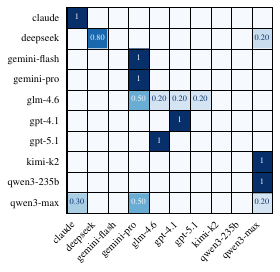}
        \caption{Ranks-JS}
    \end{subfigure}
    \begin{subfigure}{0.32\textwidth}
        \includegraphics[width=\textwidth]{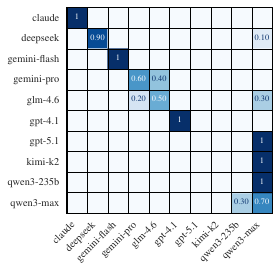}
        \caption{Entropy-JS}
    \end{subfigure}
    \begin{subfigure}{0.32\textwidth}
        \includegraphics[width=\textwidth]{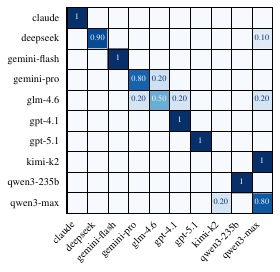}
        \caption{Entropy-Norm}
    \end{subfigure}
    \caption{Detailed normalised \oodOne confusion matrices for different techniques (in sub-captions) without selecting thresholds.
    }
    \label{fig:conf-matrix-OOD-results}
\end{figure*}

\begin{table*}[!htp]
    \centering
    \begin{tabular}{l c c c}
        \toprule[1.5pt]
        \textbf{Method} & \textbf{ID} & \textbf{\oodOne} & \textbf{\oodTwo} \\
        \midrule
        \textbf{\rank} & {$0.13 \pm 0.04$} & {$0.13 \pm 0.03$} & {$0.39 \pm 0.19$} \\
        \textbf{\entropy} & {$0.16 \pm 0.03$} & {$0.18 \pm 0.02$} & {$0.15 \pm 0.12$} \\
        \textbf{\gltr} & {$0.15 \pm 0.01$} & {$0.15 \pm 0.01$} & {$0.36 \pm 0.07$} \\
        \midrule
        \textbf{\ngram} & $0.68 \pm 0.08$ & $0.56 \pm 0.10$ & $0.32 \pm 0.33$ \\
        \textbf{\bert} & $0.70 \pm 0.10$ & \underline{$0.59 \pm 0.12$} & $0.17 \pm 0.05$ \\
        \textbf{\topformer} & $0.73 \pm 0.12$ & $0.56 \pm 0.10$ & $0.38 \pm 0.05$ \\
        \textbf{\detective} & $0.74 \pm 0.06$ & $0.56 \pm 0.15$ & $0.45 \pm 0.15$ \\
        \midrule
        \textbf{\ourmethod} \\
        \ \ Ranks-JS & $0.59 \pm 0.13$ & $0.44 \pm 0.11$ & $0.19 \pm 0.03$ \\
        \ \ Entropy-JS & \underline{$0.76 \pm 0.08$} & $0.52 \pm 0.03$ & $\bm{0.54 \pm 0.15}$ \\
        \ \ Entropy-Norm & $\bm{0.81 \pm 0.10}$ & $\bm{0.66 \pm 0.06}$ & \underline{$0.44 \pm 0.09$} \\
        \bottomrule[1.5pt]
    \end{tabular}
    \caption{Results on \ourdataset for all authors (including both \lessprolific and \moreprolific), complementing \reftab{tab:main-results}.}
    \label{tab:all_authors-main-results}
\end{table*}

\begin{table*}[!htp]
    \centering
    \resizebox{0.99\textwidth}{!}{
    \add{
    \begin{tabular}{l cccc cccc}
        \toprule[1.5pt]
        \multirow{2}{*}{\textbf{Method}} & \multicolumn{4}{c}{\textbf{ID}} & \multicolumn{4}{c}{\textbf{\oodOne}} \\
        \cmidrule(lr){2-5} \cmidrule(lr){6-9}
         & {Top-$1$} & {Top-$2$} & {Top-$3$} & {Top-$5$} & {Top-$1$} & {Top-$2$} & {Top-$3$} & {Top-$5$} \\
        \midrule
        \textbf{\rank} & $0.33 \pm 0.04$ & $0.46 \pm 0.04$ & $0.59 \pm 0.05$ & $0.81 \pm 0.05$ & $0.33 \pm 0.03$ & $0.51 \pm 0.05$ & $0.62 \pm 0.04$ & $0.86 \pm 0.02$ \\
        \textbf{\entropy} & $0.42 \pm 0.03$ & $0.52 \pm 0.00$ & $0.69 \pm 0.08$ & $0.86 \pm 0.01$ & $0.44 \pm 0.03$ & $0.47 \pm 0.01$ & $0.75 \pm 0.07$ & $0.86 \pm 0.01$ \\
        \textbf{\gltr} & $0.41 \pm 0.07$ & $0.55 \pm 0.03$ & $0.67 \pm 0.09$ & $0.82 \pm 0.06$ & $0.43 \pm 0.08$ & $0.52 \pm 0.02$ & $0.70 \pm 0.03$ & $0.85 \pm 0.04$ \\
        \midrule
        \textbf{\ngram} & $0.91 \pm 0.02$ & $0.97 \pm 0.00$ & $0.97 \pm 0.00$ & $0.97 \pm 0.00$ & \underline{$0.87 \pm 0.04$} & \bm{$0.94 \pm 0.02$} & $0.95 \pm 0.01$ & $0.97 \pm 0.00$ \\
        \textbf{\bert} & $0.92 \pm 0.05$ & $0.96 \pm 0.01$ & $0.97 \pm 0.02$ & $0.98 \pm 0.03$ & $0.85 \pm 0.05$ & $0.92 \pm 0.02$ & $0.97 \pm 0.04$ & $0.98 \pm 0.03$ \\
        \textbf{\topformer} & $0.94 \pm 0.02$ & $0.97 \pm 0.00$ & $0.97 \pm 0.00$ & $0.99 \pm 0.01$ & $0.86 \pm 0.04$ & $0.96 \pm 0.03$ & $0.97 \pm 0.02$ & $0.99 \pm 0.01$ \\
        \textbf{\detective} & \underline{$0.93 \pm 0.02$} & $0.96 \pm 0.02$ & \underline{$0.98 \pm 0.02$} & \underline{$0.99 \pm 0.01$} & \bm{$0.89 \pm 0.02$} & \bm{$0.94 \pm 0.02$} & $0.95 \pm 0.01$ & $0.96 \pm 0.01$ \\
        \midrule
        \textbf{\ourmethod} \\
        \ \ Ranks-JS & $0.80 \pm 0.09$ & $0.95 \pm 0.02$ & $0.97 \pm 0.00$ & $0.97 \pm 0.00$ & $0.69 \pm 0.09$ & $0.79 \pm 0.04$ & $0.85 \pm 0.03$ & \underline{$0.98 \pm 0.01$} \\
        \ \ Entropy-JS & $0.90 \pm 0.05$ & \bm{$0.98 \pm 0.01$} & \bm{$0.98 \pm 0.01$} & \underline{$0.99 \pm 0.01$} & $0.76 \pm 0.05$ & $0.90 \pm 0.03$ & $0.93 \pm 0.03$ & \bm{$0.99 \pm 0.01$} \\
        \ \ Entropy-Norm & \bm{$0.94 \pm 0.03$} & \underline{$0.97 \pm 0.00$} & $0.97 \pm 0.00$ & \bm{$1.00 \pm 0.00$} & $0.83 \pm 0.04$ & \underline{$0.93 \pm 0.01$} & \bm{$0.97 \pm 0.00$} & \bm{$0.99 \pm 0.01$} \\
        \bottomrule[1.5pt]
    \end{tabular}
    }
    }
    \caption{\add{Top-$k$ accuracy results on \ourdataset for all authors, complementing \reftab{tab:main-results}~and~\reftab{tab:all_authors-main-results}.}}
    \label{tab:all_authors-top-k-results}
\end{table*}

\begin{table*}[!htp]
    \centering
    \begin{tabular}{lcc|cc}
        \toprule[1.5pt]
        \multirow{2}{*}{\textbf{Method}} & \multicolumn{2}{c|}{\textbf{Low Resource}} & \multicolumn{2}{c}{\textbf{High Resource}} \\
        \cmidrule(lr){2-3} \cmidrule(lr){4-5}
            & {ID} & {\oodOne} & {ID} & {\oodOne} \\
        \midrule
        \ngram & $0.47 \pm 0.09$ & $0.20 \pm 0.16$ & {$0.95 \pm 0.01$} & \underline{$0.92 \pm 0.04$} \\
        \bert & $0.51 \pm 0.23$ & $0.24 \pm 0.17$ & {$0.95 \pm 0.04$} & $0.91 \pm 0.06$ \\
        \topformer & \bm{$0.69 \pm 0.06$} & $0.31 \pm 0.08$ & \underline{$0.96 \pm 0.04$} & $0.91 \pm 0.05$ \\
        \detective & {$0.58 \pm 0.14$} & {$0.31 \pm 0.08$} & \underline{$0.96 \pm 0.04$} & $\bm{0.93 \pm 0.04}$ \\
        \midrule
        \textbf{\ourmethod (Ours)} & \\
        \ \ Ranks-JS & $0.39 \pm 0.25$ & $0.21 \pm 0.19$ & $0.81 \pm 0.04$ & $0.73 \pm 0.13$ \\
        \ \ Entropy-JS & {$0.64 \pm 0.20$} & \underline{$0.34 \pm 0.12$} & $0.91 \pm 0.05$ & $0.81 \pm 0.06$ \\
        \ \ Entropy-Norm & $\underline{0.67 \pm 0.19}$ & $\bm{0.56 \pm 0.19}$ & $\bm{0.96 \pm 0.02}$ & $0.85 \pm 0.05$ \\
        \bottomrule[1.5pt]
    \end{tabular}
    \caption{Without thresholding results for attribution techniques on \ourdataset, showing ID and \oodOne for Low and High resource settings, complementing \reftab{tab:main-results}. \textbf{Bold} denotes the best result and \underline{underline} the second best.}
    \label{table:w/o-thresholding-results}
\end{table*}

\begin{table*}[!htp]
    \centering
    \begin{tabular}{lcc|cc}
        \toprule[1.5pt]
        \multirow{2}{*}{\textbf{Method}} & \multicolumn{2}{c|}{\textbf{Low Resource}} & \multicolumn{2}{c}{\textbf{High Resource}} \\
        \cmidrule(lr){2-3} \cmidrule(lr){4-5}
         & Reject & Family & Reject & Family \\
        \midrule
        \ngram & $0.36 \pm 0.31$ & $0.19 \pm 0.26$ & $0.28 \pm 0.35$ & $0.03 \pm 0.04$ \\
        \bert & $0.04 \pm 0.06$ & $0.55 \pm 0.10$ & $0.29 \pm 0.05$ & $0.09 \pm 0.08$ \\
        \topformer & $0.29 \pm 0.17$ & $0.55 \pm 0.10$ & $0.46 \pm 0.27$ & $0.15 \pm 0.13$ \\
        \detective & $0.40 \pm 0.20$ & $0.49 \pm 0.18$ & $0.49 \pm 0.22$ & $0.15 \pm 0.11$ \\
        \midrule
        \textbf{\ourmethod (Ours)} & \\
        \ \ Ranks-JS & $0.18 \pm 0.03$ & $0.53 \pm 0.22$ & $0.20 \pm 0.08$ & $0.32 \pm 0.12$ \\
        \ \ Entropy-JS & $0.49 \pm 0.14$ & $0.11 \pm 0.16$ & $0.58 \pm 0.20$ & $0.19 \pm 0.07$ \\
        \ \ Entropy-Norm & $0.51 \pm 0.17$ & $0.11 \pm 0.16$ & $0.37 \pm 0.03$ & $0.25 \pm 0.11$ \\
        \bottomrule[1.5pt]
    \end{tabular}
    \caption{Variant of \oodTwo evaluation. `Family' column: if the predicted class for the unseen LLM belongs to the same family, we treat as a correct prediction (instead of rejecting: `Reject' column).}
    \label{tab:family-OOD-results}
\end{table*}

\begin{table*}[!htp]
    \centering

    \subfloat[Rank-based.]{
    \resizebox{0.99\textwidth}{!}{
    \begin{tabular}{lcccccc}
        \toprule[1.5pt]
        \multirow{2}{*}{\textbf{Model}}  & \multicolumn{3}{c}{\textbf{Low Resource}} & \multicolumn{3}{c}{\textbf{High Resource}} \\
        \cmidrule(lr){2-4} \cmidrule(lr){5-7}
            & ID & \oodOne & \oodTwo 
            & ID & \oodOne & \oodTwo \\
        \midrule
        \textbf{\gpt} 
            & $0.40 \pm 0.24$ & $0.22 \pm 0.19$ & $0.18 \pm 0.03$
            & $0.78 \pm 0.06$ & $0.66 \pm 0.12$ & $0.20 \pm 0.08$ \\
        \textbf{\olmo} 
            & $0.22 \pm 0.14$ & $0.32 \pm 0.22$ & $0.11 \pm 0.08$
            & $0.74 \pm 0.05$ & $0.56 \pm 0.08$ & $0.18 \pm 0.05$ \\
        \textbf{\gemma} 
            & $0.33 \pm 0.22$ & $0.27 \pm 0.04$ & $0.11 \pm 0.16$
            & $0.79 \pm 0.06$ & $0.63 \pm 0.08$ & $0.15 \pm 0.03$ \\
        \midrule
    \end{tabular}}}
    \vspace{2mm}

    \subfloat[Entropy-based (JS Dist.).]{
    \resizebox{0.99\textwidth}{!}{
    \begin{tabular}{lcccccc}
        \toprule[1.5pt]
        \multirow{2}{*}{\textbf{Model}}  & \multicolumn{3}{c}{\textbf{Low Resource}} & \multicolumn{3}{c}{\textbf{High Resource}} \\
        \cmidrule(lr){2-4} \cmidrule(lr){5-7}
            & ID & \oodOne & \oodTwo 
            & ID & \oodOne & \oodTwo \\
        \midrule
        \textbf{\gpt} 
            & $0.60 \pm 0.16$ & $0.23 \pm 0.12$ & $0.49 \pm 0.14$
            & $0.91 \pm 0.06$ & $0.82 \pm 0.08$ & $0.58 \pm 0.20$ \\
        \textbf{\olmo} 
            & $0.71 \pm 0.08$ & $0.24 \pm 0.11$ & $0.29 \pm 0.08$
            & $0.95 \pm 0.02$ & $0.75 \pm 0.12$ & $0.47 \pm 0.13$ \\
        \textbf{\gemma} 
            & $0.51 \pm 0.08$ & $0.10 \pm 0.08$ & $0.42 \pm 0.08$
            & $0.89 \pm 0.06$ & $0.80 \pm 0.09$ & $0.54 \pm 0.18$ \\
        \midrule
    \end{tabular}}}
    \vspace{2mm}

    \subfloat[Entropy-based (norm-mean).]{
    \resizebox{0.99\textwidth}{!}{
    \begin{tabular}{lcccccc}
        \toprule[1.5pt]
        \multirow{2}{*}{\textbf{Model}} & \multicolumn{3}{c}{\textbf{Low Resource}} & \multicolumn{3}{c}{\textbf{High Resource}} \\
        \cmidrule(lr){2-4} \cmidrule(lr){5-7}
         & ID & \oodOne & \oodTwo 
         & ID & \oodOne & \oodTwo \\
        \midrule
        \textbf{\gpt} 
            & $0.67 \pm 0.19$ & $0.49 \pm 0.11$ & $0.51 \pm 0.17$
            & $0.96 \pm 0.02$ & $0.82 \pm 0.10$ & $0.37 \pm 0.03$ \\
        \textbf{\olmo} 
            & $0.70 \pm 0.08$ & $0.34 \pm 0.10$ & $0.29 \pm 0.11$
            & $0.92 \pm 0.04$ & $0.81 \pm 0.09$ & $0.54 \pm 0.09$ \\
        \textbf{\gemma} 
            & $0.33 \pm 0.09$ & $0.20 \pm 0.16$ & $0.67 \pm 0.11$
            & $0.87 \pm 0.13$ & $0.80 \pm 0.09$ & $0.69 \pm 0.09$ \\
        \midrule
    \end{tabular}}}

    \caption{\ourmethod performance for different evaluator language models. Each sub-table is for a variant of the \ourmethod fingerprint. Overall, we observe comparable performance using different evaluator language models.}
    \label{tab:eval-llms-all}
\end{table*}

\begin{table}[!htp]
    \centering
    \resizebox{0.99\columnwidth}{!}{
    \begin{tabular}{clll}
        \toprule[1.5pt]
        {\textbf{Scenario}} & \textbf{Split 1} & \textbf{Split 2} & \textbf{Split 3} \\
        \midrule 
        \multirow{5}{*}{\rotatebox{90}{\lessprolific}}
            & \geminiFlash & \geminiFlash & \geminiPro \\
            & \glm & \claude  & \geminiFlash \\
            & \deepseek & \kimi & \qwen \\
            & \gptFive & \gptFive & \qwenMax \\
            & \qwen & \qwen & \deepseek \\
        \midrule
        \multirow{5}{*}{\rotatebox{90}{\moreprolific}}
            & \geminiPro & \geminiPro & \glm \\
            & \claude & \deepseek & \claude \\
            & \kimi & \gptFour & \kimi \\
            & \gptFour & \glm & \gptFour \\
            & \qwenMax & \qwenMax & \gptFive \\
    \bottomrule[1.5pt]
    \end{tabular}}
    \caption{Random LLMs splits for \lessprolific and \moreprolific scenarios. These splits are released with the \ourdataset dataset.}
    \label{tab:prolificity-splits}
\end{table}

\begin{table}[!htp]
    \centering
    \begin{tabular}{l lll}
        \toprule[1.5pt]
        \textbf{Method} & \textbf{Split 1} & \textbf{Split 2} & \textbf{Split 3} \\
        \midrule 
        \rank & 0.30 & 0.30 & 0.35 \\
        \entropy & 0.30 & 0.25 & 0.25 \\
        \gltr &  0.30 & 0.30 & 0.25 \\
        \midrule
        \midrule
        \ngram & 0.40 & 0.75 & 0.40 \\
        \bert & 0.50 & 0.50 & 0.50 \\
        \topformer & 0.60 & 0.60 & 0.70 \\
        \detective & 0.50 & 0.90 & 0.80 \\
        \midrule
        \midrule
        \ourmethod \\
        \ \ \gpt \\
        \ \ \ \ Ranks & 0.92 & 0.92 & 0.92 \\
        \ \ \ \ Entropy-JS & 0.95 & 0.93 & 0.94 \\
        \ \ \ \ Entropy-Norm & -4.0 & -4.0 & -4.5 \\
        \midrule
        \ \ \olmo \\
        \ \ \ \ Ranks & 0.92 & 0.92 & 0.92 \\
        \ \ \ \ Entropy-JS & 0.94 & 0.93 & 0.95 \\
        \ \ \ \ Entropy-Norm & -3.5 & -4.0 & -3.5 \\
        \midrule
        \ \ \gemma \\
        \ \ \ \ Ranks & 0.92 & 0.91 & 0.92 \\
        \ \ \ \ Entropy-JS & 0.94 & 0.94 & 0.95 \\
        \ \ \ \ Entropy-Norm & -2.75 & -2.50 & -3.25 \\
        \midrule
    \end{tabular}
    \caption{Threshold (tuned as per development set) for different splits (from \reftab{tab:prolificity-splits}) and attribution methods used in this work. `Norm' represents the norm-mean metric (as angle) as defined \refapp{sec:sim-metrics} and for this lower is better (hence negative values reported). All norm-mean values are multiplied by 100 for presentation. }
    \label{tab:thresholds-config}
\end{table}

\clearpage
\begin{figure*}[!htp]
    \centering
    \includegraphics[width=0.99\linewidth]{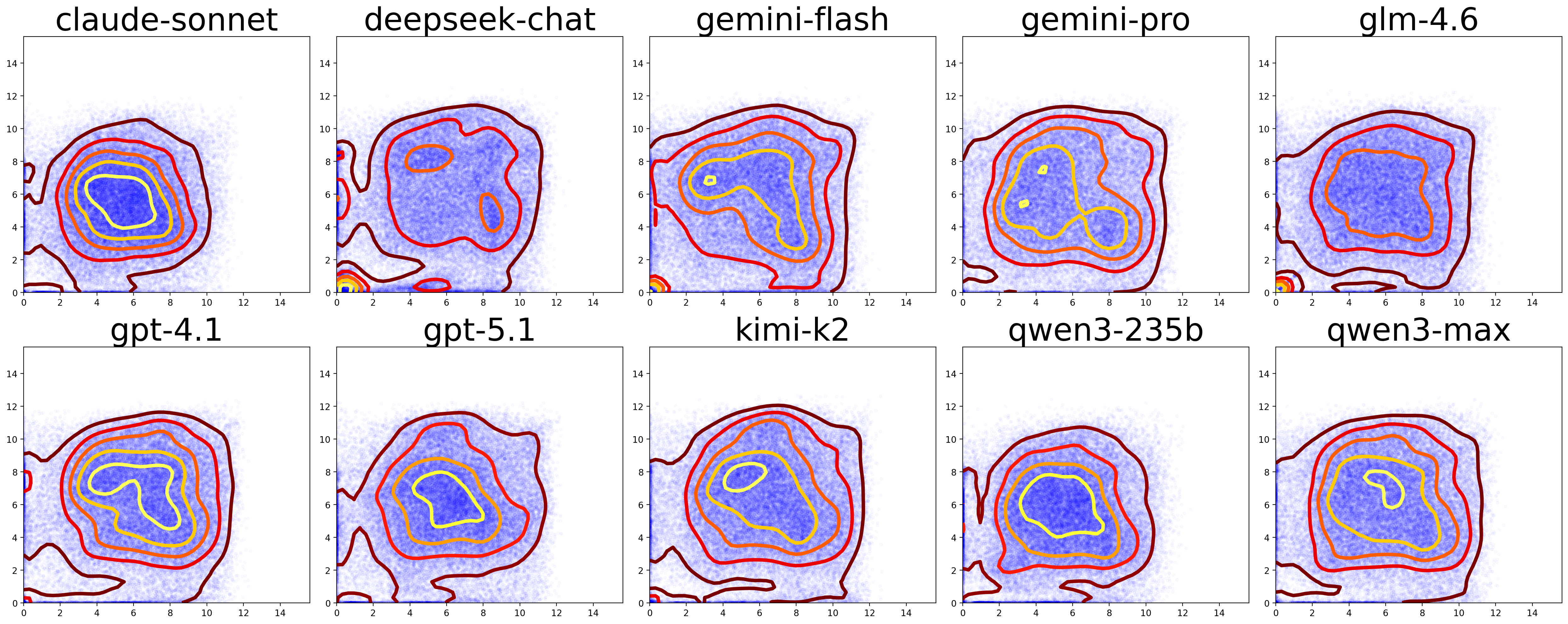}
    \caption{\ourmethod Entropy-based reference fingerprints learned for different LLMs (in subcaption) in \ourdataset. \add{Each axis represents the entropy of consecutive token transitions estimated by an independent language model; denser regions indicate more frequent transition types. We note distinct structural patterns for each LLM, suggesting they capture their generation styles.}
    }
    \label{fig:entropy-ref-fingerprints}
\end{figure*}

\begin{figure*}
    \centering
    \includegraphics[width=0.99\linewidth]{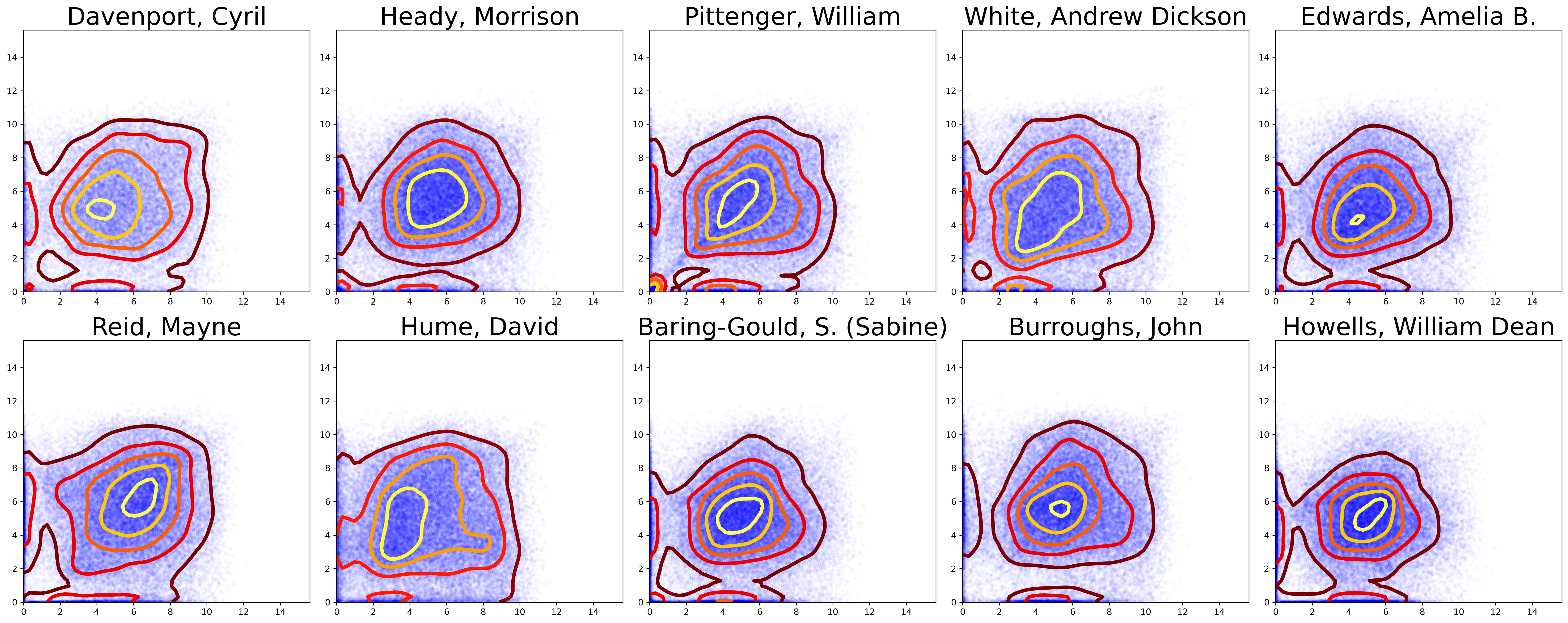}
    \caption{\add{\ourmethod Entropy-based reference fingerprints (which are very different from LLM counterparts in \reffig{fig:entropy-ref-fingerprints}) learned for different human authors (in subcaption; sourced from Project Gutenberg) used for unseen authors ablation study \refsec{sec:unseen-human-results}.}}
    \label{fig:unseen-humans-entropy-ref-fingerprints}
\end{figure*}

\begin{figure*}[!htp]
    \centering
    \includegraphics[width=0.99\linewidth]{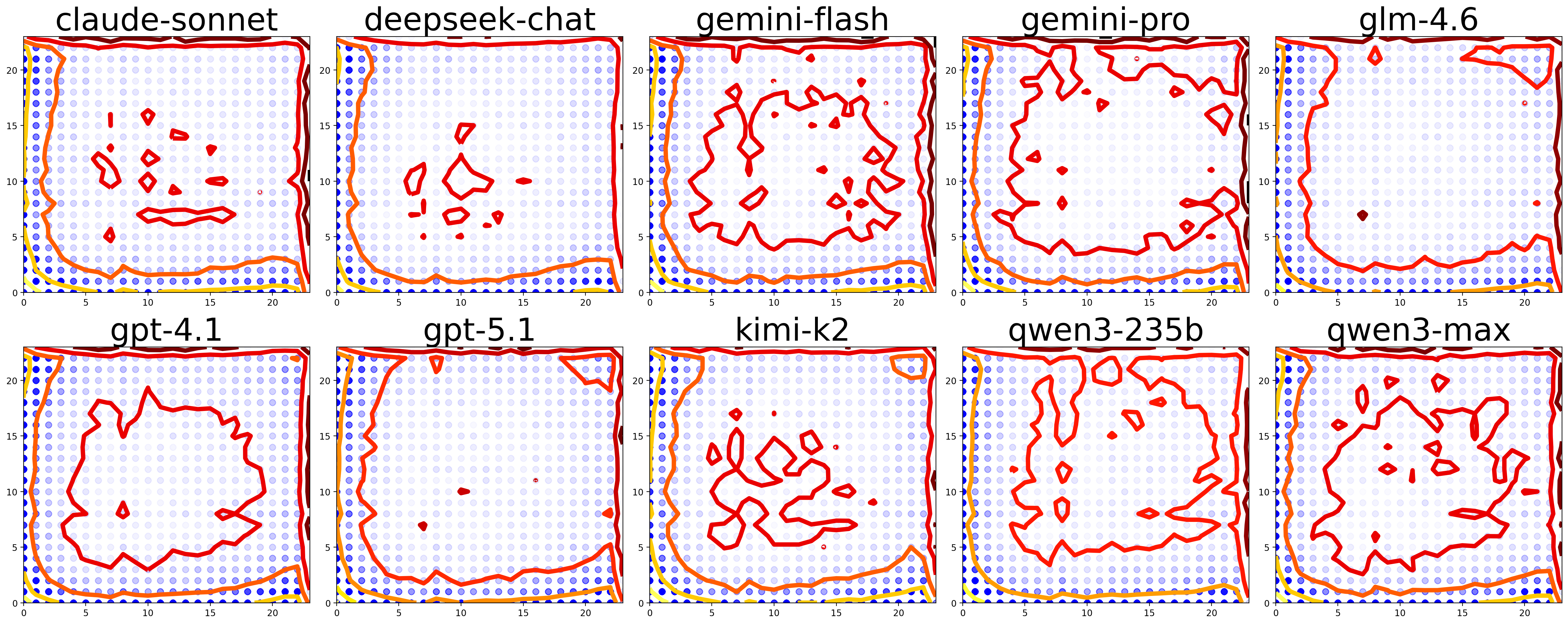}
    \caption{\ourmethod Rank-based reference fingerprints (in log-scale) learned for different LLMs (in subcaption). Although not as pronounced as Entropy-based (as seen in \reffig{fig:entropy-ref-fingerprints}), some structural patterns are there for each LLM. Note: Rank-based fingerprints have a grid structure due to the discrete nature of rank compared to continuous (scatter plot) entropy values.}
    \label{fig:ranks-ref-fingerprints}
\end{figure*}

\clearpage
\begin{figure*}[!h]
    \centering
    \begin{tcblisting}{
        listing only,                     %
        titlerule=0pt,  
        colbacktitle=purple!30,
        coltitle=purple!80!black,
        fonttitle=\bfseries,
        colback=purple!10,
        colframe=purple!50!black,
        arc=2mm,                  %
        boxrule=0.8pt,            %
        left=1pt, right=1pt, top=1pt, bottom=1pt,
    }
Below is the beginning part of a story:

---

{}

---

We are going over segments of a story sequentially to gradually update one comprehensive summary of the entire plot. Write a summary for the excerpt provided above, make sure to include vital information related to key events, backgrounds, settings, characters, their objectives, and motivations. You must briefly introduce characters, places, and other major elements if they are being mentioned for the first time in the summary. The story may feature non-linear narratives, flashbacks, switches between alternate worlds or viewpoints, etc. Therefore, you should organize the summary so it presents a consistent and chronological narrative. Despite this step-by-step process of updating the summary, you need to create a summary that seems as though it is written in one go. The summary could include multiple paragraphs.

Summary:
    \end{tcblisting}
    \vspace{-1.2em}
    \caption{Generate Initial Summary Prompt, adapted from \citep{chang2024booookscore}.}
\end{figure*}

\begin{figure*}[!h]
    \centering
    \begin{tcblisting}{
        listing only,                     %
        titlerule=0pt,  
        colbacktitle=purple!30,
        coltitle=purple!80!black,
        fonttitle=\bfseries,
        colback=purple!10,
        colframe=purple!50!black,
        arc=2mm,                  %
        boxrule=0.8pt,            %
        left=1pt, right=1pt, top=1pt, bottom=1pt,
    }
Below is a segment from a story:

---

{}

---

Below is a summary of the story up until this point:

---

{}

---

We are going over segments of a story sequentially to gradually update one comprehensive summary of the entire plot. You are required to update the summary to incorporate any new vital information in the current excerpt. This information may relate to key events, backgrounds, settings, characters, their objectives, and motivations. You must briefly introduce characters, places, and other major elements if they are being mentioned for the first time in the summary. The story may feature non-linear narratives, flashbacks, switches between alternate worlds or viewpoints, etc. Therefore, you should organize the summary so it presents a consistent and chronological narrative. Despite this step-by-step process of updating the summary, you need to create a summary that seems as though it is written in one go. The updated summary could include multiple paragraphs.

Updated summary:
    \end{tcblisting}
    \vspace{-1.2em}
    \caption{Generate Intermediate Summary Prompt, adapted from \citep{chang2024booookscore}.}
    \label{prompt:summarise}
\end{figure*}

\begin{figure*}[!h]
    \centering
    \begin{tcblisting}{
        listing only,                     %
        titlerule=0pt,  
        colbacktitle=purple!30,
        coltitle=purple!80!black,
        fonttitle=\bfseries,
        colback=purple!10,
        colframe=purple!50!black,
        arc=2mm,                  %
        boxrule=0.8pt,            %
        left=1pt, right=1pt, top=1pt, bottom=1pt,
    }
Below is a summary of part of a story:

---

{}

---

Currently, this summary contains {} words. Your task is to condense it to less than {} words. The condensed summary should remain clear, overarching, and fluid while being brief. Whenever feasible, maintain details about key events, backgrounds, settings, characters, their objectives, and motivations - but express these elements more succinctly. Make sure to provide a brief introduction to characters, places, and other major components during their first mention in the condensed summary. Remove insignificant details that do not add much to the overall story line. The story may feature non-linear narratives, flashbacks, switches between alternate worlds or viewpoints, etc. Therefore, you should organize the summary so it presents a consistent and chronological narrative.

Condensed summary (to be within {} words):
    \end{tcblisting}
    \vspace{-1.2em}
    \caption{Summary Compression Prompt, adapted from \citep{chang2024booookscore}.}
    \label{prompt:compress-summary}
\end{figure*}

\begin{figure*}[!h]
    \centering
    \begin{tcblisting}{
        listing only,                     %
        titlerule=0pt,  
        colbacktitle=orange!30,
        coltitle=orange!80!black,
        fonttitle=\bfseries,
        colback=orange!10,
        colframe=orange!50!black,
        arc=2mm,                  %
        boxrule=0.8pt,            %
        left=1pt, right=1pt, top=1pt, bottom=1pt,
    }
Generate a detailed outline for a {}-word novel in your own style based on these genres: {} set in the {} century. Include word counts and major plot points.

Below you are given some snippets of similar novels just for reference:

SNIPPET 1: {}
    .
    .
    .
SNIPPET 6: {}

Please insert the generated text between <START>...<END> tags. Any other message or metadata outside these tags.
    \end{tcblisting}
    \vspace{-1.2em}
    \caption{\Doc Outline Generation Prompt.}
    \label{prompt:outline}
\end{figure*}

\begin{figure*}[!h]
    \centering
    \begin{tcblisting}{   
        listing only,                     %
        titlerule=0pt,  
        colbacktitle=orange!30,
        coltitle=orange!80!black,
        fonttitle=\bfseries,
        colback=orange!10,
        colframe=orange!50!black,
        arc=2mm,                  %
        boxrule=0.8pt,            %
        left=1pt, right=1pt, top=1pt, bottom=1pt,
    }
GOAL: Write a {} words novel in your own style based on these genres: {}. The novel is set in the {} century. We are generating the complete novel sequentially and in parts. 

TASK: Begin writing the novel by generating the initial segment (~{} words). Expand the narrative comprehensively, including all relevant details, while strictly adhering to the word limit specified in the outline. Do not rush through any milestone; it is acceptable to split the content across multiple parts if needed to respect the word limits. Please enclose the generated text strictly between <START> and <END> tags. Any other message or metadata outside these tags. 

Please read the following text carefully:
Follow the given novel outline strictly (including word counts and major plot points).

OUTLINE:
{}

Below you are given some snippets of similar novels for reference:
SNIPPET 1: {}
    .
    .
    .
SNIPPET 6: {}
    \end{tcblisting}
    \vspace{-1.2em}
    \caption{\Doc Generation Start Prompt.}
    \label{prompt:start-novel-segment}
\end{figure*}

\begin{figure*}[!h]
    \centering
    \begin{tcblisting}{  
        listing only,                     %
        titlerule=0pt,  
        colbacktitle=orange!30,
        coltitle=orange!80!black,
        fonttitle=\bfseries,
        colback=orange!10,
        colframe=orange!50!black,
        arc=2mm,                  %
        boxrule=0.8pt,            %
        left=1pt, right=1pt, top=1pt, bottom=1pt,
    }
GOAL: Write a {} words novel in your own style based on these genres: {}. The novel is set in the {} century. We are generating the complete novel sequentially and in parts. 

TASK: You are continuing the novel generation process. Generate the next segment (~{} words) of the novel. Expand the narrative comprehensively, including all relevant details, while strictly adhering to the word limit specified in the outline. Do not rush through any milestone; it is acceptable to split the content across multiple parts if needed to respect the word limits. Please enclose the generated text strictly between <START> and <END> tags. If novel is complete, add "END OF NOVEL" at the end of the generated text. Any other message or metadata outside these tags.

INSTRUCTIONS:
1. Follow the novel outline strictly, including word counts and major plot points.
2. Maintain full continuity with the previous segment and the overall summary generated so far.
3. Use the previous segments and summary as reference for style, tone, and story consistency.
4. Draw inspiration from the provided snippet texts for style, but do not copy directly.

OUTLINE:
{}

ENDING OF PREVIOUS SEGMENT:
{}

SUMMARY OF NOVEL TILL NOW:
{}

REFERENCE SNIPPETS:
SNIPPET 1: {}
    .
    .
    .
SNIPPET 6: {}
    \end{tcblisting}
    \vspace{-1.2em}
    \caption{\Doc Generation Incremental Prompt.}
    \label{prompt:continue-novel-segment}
\end{figure*}

\end{document}

%% file: def.tex
\usepackage{stmaryrd}
\usepackage{amsfonts}
\usepackage{amssymb}
\usepackage{amsmath}
\usepackage{mathtools}
\usepackage{bbm}
\usepackage{xspace}
\usepackage{todonotes}
\usepackage{adjustbox}
\usepackage{multicol}
\usepackage{spverbatim}
\usepackage{pifont}

\usepackage{algorithm}
\usepackage{algpseudocode}
\usepackage[most]{tcolorbox}
\tcbuselibrary{listings,breakable,skins}
\usepackage{listings}
\usepackage{makecell} 
\usepackage{fontawesome5}
\usepackage{arydshln} %

\usepackage{arydshln}
\usepackage{tabularx,booktabs,multirow}  %
\usepackage{tablefootnote}
\usepackage{subcaption}
\usepackage{enumerate}
\usepackage{siunitx}
\usepackage{soul} %
\usepackage{pgfplots}
\usepackage[table]{xcolor}
\usepackage{enumitem}

\input{math_commands}

\newcommand{\add}[1]{{{#1}}}
\newcommand{\addTwo}[1]{{{#1}}}

\newcommand{\addThree}[1]{#1}

\newcommand{\cmark}{\textcolor{green!70!black}{\ding{51}}}   %
\newcommand{\xmark}{\textcolor{red}{\ding{55}}}              %

\newcommand{\ngram}{\textsc{n-gram}\xspace}
\newcommand{\bert}{\textsc{Bert-AA}\xspace}
\newcommand{\topformer}{\textsc{TopFormer}\xspace}
\newcommand{\detective}{\textsc{DeTeCtive}\xspace}
\newcommand{\rank}{\textsc{Rank}\xspace}
\newcommand{\entropy}{\textsc{Entropy}\xspace}
\newcommand{\gltr}{\textsc{GLTR}\xspace}

\newcommand{\fscore}{$\text{macro-F}_1$\xspace}

\newcommand{\oodOne}{OOD-Domain\xspace}
\newcommand{\oodTwo}{OOD-Author\xspace}

\newcommand{\doc}{book\xspace}
\newcommand{\Doc}{Book\xspace}

\newcommand{\lessprolific}{low-resource\xspace}
\newcommand{\moreprolific}{high-resource\xspace}

\newcommand{\ourmethod}{\textsc{TRACE}\xspace}

\newcommand{\ourdataset}{\textsc{GhostWriteBench}\xspace}

\newcommand{\glm}{\texttt{glm-4.6}\xspace}
\newcommand{\qwen}{\texttt{qwen3-235b}\xspace}
\newcommand{\qwenMax}{\texttt{qwen3-max}\xspace}
\newcommand{\geminiPro}{\texttt{gemini-pro}\xspace}
\newcommand{\geminiFlash}{\texttt{gemini-flash}\xspace}
\newcommand{\gpt}{\texttt{GPT-2}\xspace}
\newcommand{\gptFour}{\texttt{gpt-4.1}\xspace}
\newcommand{\gptFive}{\texttt{gpt-5.1}\xspace}
\newcommand{\claude}{\texttt{claude-sonnet}\xspace}
\newcommand{\deepseek}{\texttt{deepseek-chat}\xspace}
\newcommand{\kimi}{\texttt{kimi-k2}\xspace}

\newcommand{\olmo}{\texttt{OLMo}\xspace}
\newcommand{\gemma}{\texttt{Gemma}\xspace}

\newcommand{\selfBleu}{Self-\textsc{BLEU}\xspace}

\newcommand{\refappalg}[1]{Appendix~Algorithm~\ref{#1}}

\newcommand{\refapp}[1]{Appendix~\ref{#1}}
\newcommand{\refapptab}[1]{Appendix~Table~\ref{#1}}
\newcommand{\refappfig}[1]{Appendix~Figure~\ref{#1}}

\newcommand{\reffig}[1]{Figure~\ref{#1}}
\newcommand{\refsec}[1]{Section~\ref{#1}}
\newcommand{\reftab}[1]{Table~\ref{#1}}

\def\eg{{e.g.,}\xspace}
\def\ie{{i.e.,}\xspace}
\def\versus{{vs.}\xspace}

\def\aka{{a.k.a}\xspace}
\def\etc{{etc.}\xspace}

%% file: math_commands.tex
\usepackage{amsmath,amsfonts,bm}

\def\eqref#1{equation~\ref{#1}}

\def\1{\bm{1}}

\DeclareMathAlphabet{\mathsfit}{\encodingdefault}{\sfdefault}{m}{sl}
\SetMathAlphabet{\mathsfit}{bold}{\encodingdefault}{\sfdefault}{bx}{n}